\documentclass{article}
\usepackage[utf8]{inputenc}
\usepackage{main}
\usepackage{microtype}
\usepackage{graphicx}
\usepackage{subfig}
\usepackage{times}
\usepackage{latexsym}
\usepackage{amsmath}
\usepackage{amssymb}
\usepackage{float}
\usepackage{placeins}
\usepackage{footnote}
\usepackage{enumitem}
\usepackage{bm}
\usepackage{arydshln}
\usepackage{booktabs}
\usepackage{multicol}
\usepackage{multirow}
\usepackage{color}
\usepackage{xcolor}   

\definecolor{opensiiBlue}{RGB}{127,126,236}
\definecolor{myblue}{rgb}{0.9, 0.1, 0.94}
\definecolor{mygreen}{rgb}{0.604, 0.522, 1.000}
\definecolor{myyellow}{rgb}{0.68, 0.6, 0.1}
\definecolor{fancygreen}{rgb}{0.33, 0.68, 0.20}
\definecolor{salmon}{rgb}{0.94, 0.52, 0.49}
\definecolor{tablegreen}{rgb}{0.82, 0.94, 0.75}
\definecolor{tableblue}{rgb}{0.81, 0.90, 0.94}
\definecolor{tablered}{rgb}{0.97, 0.85, 0.85}
\definecolor{tableorange}{rgb}{0.96, 0.85, 0.81}
\definecolor{academicblue}{rgb}{0.95, 0.97, 1.0}
\definecolor{academicpink}{rgb}{1.0, 0.95, 0.95}
\definecolor{ForestGreen}{rgb}{0.133, 0.545, 0.133}
\usepackage{colortbl}
\usepackage{bbding}
\usepackage{makecell}
\usepackage{mathtools}
\usepackage{imakeidx}
\usepackage{longtable}
\usepackage{wrapfig}
\makeindex
\usepackage{arydshln}
\usepackage{lipsum}
\usepackage{natbib}
\usepackage[toc]{multitoc}
\usepackage[edges]{forest}
\usepackage[normalem]{ulem}
\definecolor{mydarkblue}{rgb}{0,0.08,0.45}
\usepackage[colorlinks=true,linkcolor=mydarkblue,citecolor=mydarkblue,filecolor=mydarkblue,urlcolor=mydarkblue]{hyperref}
\usepackage{caption}
\usepackage{CJKutf8}
\usepackage{awesomebox} 
\usepackage{bbding}
\usepackage[most]{tcolorbox}

\usepackage{microtype}
\usepackage{hyperref}
\usepackage{url}
\usepackage{booktabs}
\usepackage{array}
\usepackage{amsmath}
\usepackage{graphicx}
\usepackage{subcaption}
\usepackage{multirow}
\usepackage{wrapfig}
\usepackage[table]{xcolor}
\usepackage{colortbl}
\usepackage{tcolorbox}
\usepackage{enumitem}
\usepackage{svg}
\usepackage{caption}
\tcbuselibrary{breakable,skins}
\tcbset{
  rttpromptbox/.style={
    enhanced,
    rounded corners,
    breakable,
    fonttitle=\bfseries\color{white},
    coltitle=white,
  },
}
\usepackage{listings}

\definecolor{deltagreen}{rgb}{0.0, 0.55, 0.0}
\definecolor{deltared}{rgb}{0.85, 0.0, 0.0}
\definecolor{rowblue}{rgb}{0.88, 0.95, 1.0}
\newcommand{\up}[1]{\textcolor{deltagreen}{$\uparrow$ #1}}
\newcommand{\dn}[1]{\textcolor{deltared}{$\downarrow$ #1}}
\newcommand{\nc}[1]{\textcolor{gray}{$\uparrow$ #1}}

\usepackage{lineno}

\definecolor{darkblue}{rgb}{0, 0, 0.5}
\hypersetup{colorlinks=true, citecolor=darkblue, linkcolor=darkblue, urlcolor=darkblue}
\usepackage[tikz]{bclogo}
\usepackage[framemethod=tikz]{mdframed}
\definecolor{bgblue}{RGB}{245,243,253}
\definecolor{ttblue}{RGB}{91,194,224}

\mdfdefinestyle{mystyle}{%
  rightline=true,
  innerleftmargin=10,
  innerrightmargin=10,
  outerlinewidth=3pt,
  topline=false,
  rightline=true,
  bottomline=false,
  skipabove=\topsep,
  skipbelow=\topsep
}

\newtcolorbox{myboxi}[1][]{
  breakable,
  title=#1,
  colback=red!5,
  colbacktitle=red!5,
  coltitle=black,
  fonttitle=\bfseries,
  bottomrule=0pt,
  toprule=0pt,
  leftrule=2pt,
  rightrule=2pt,
  titlerule=0pt,
  arc=0pt,
  outer arc=0pt,
  colframe=red,
}

\newtcolorbox{myboxnote}[1][]{
  breakable,
  title=#1,
  colback=orange!0,
  colbacktitle=orange!0,
  coltitle=black,
  fonttitle=\bfseries,
  bottomrule=0pt,
  toprule=0pt,
  leftrule=2pt,
  rightrule=2pt,
  titlerule=0pt,
  arc=0pt,
  outer arc=0pt,
  colframe=orange,
}

\newtcolorbox{myboxii}[1][]{
  breakable,
  freelance,
  title=#1,
  colback=white,
  colbacktitle=white,
  coltitle=black,
  fonttitle=\bfseries,
  bottomrule=0pt,
  boxrule=0pt,
  colframe=white,
  overlay unbroken and first={
  \draw[red!75!black,line width=3pt]
    ([xshift=5pt]frame.north west) -- 
    (frame.north west) -- 
    (frame.south west);
  \draw[red!75!black,line width=3pt]
    ([xshift=-5pt]frame.north east) -- 
    (frame.north east) -- 
    (frame.south east);
  },
  overlay unbroken app={
  \draw[red!75!black,line width=3pt,line cap=rect]
    (frame.south west) -- 
    ([xshift=5pt]frame.south west);
  \draw[red!75!black,line width=3pt,line cap=rect]
    (frame.south east) -- 
    ([xshift=-5pt]frame.south east);
  },
  overlay middle and last={
  \draw[red!75!black,line width=3pt]
    (frame.north west) -- 
    (frame.south west);
  \draw[red!75!black,line width=3pt]
    (frame.north east) -- 
    (frame.south east);
  },
  overlay last app={
  \draw[red!75!black,line width=3pt,line cap=rect]
    (frame.south west) --
    ([xshift=5pt]frame.south west);
  \draw[red!75!black,line width=3pt,line cap=rect]
    (frame.south east) --
    ([xshift=-5pt]frame.south east);
  },
}

\usepackage{fancyhdr} 
\usepackage{blindtext} 
\usepackage{makecell}

\DeclareCaptionFont{black}{\color{black}}

\definecolor{opensiiBlue}{RGB}{127,126,236}
\definecolor{myblue}{rgb}{0.9, 0.1, 0.94}
\definecolor{mygreen}{rgb}{0.604, 0.522, 1.000}
\definecolor{myyellow}{rgb}{0.68, 0.6, 0.1}
\definecolor{fancygreen}{rgb}{0.33, 0.68, 0.20}
\definecolor{salmon}{rgb}{0.94, 0.52, 0.49}
\definecolor{tablegreen}{rgb}{0.82, 0.94, 0.75}
\definecolor{tableblue}{rgb}{0.81, 0.90, 0.94}
\definecolor{tablered}{rgb}{0.97, 0.85, 0.85}
\definecolor{tableorange}{rgb}{0.96, 0.85, 0.81}

\newenvironment{itemize*}%
 {\leftmargini=10pt\begin{itemize}%
  \setlength{\itemsep}{0pt}%
  \setlength{\parskip}{0pt}%
  }%
 {\end{itemize}}
\newenvironment{enumerate*}%
 {\begin{enumerate}%
  \setlength{\itemsep}{0pt}%
  \setlength{\parskip}{0pt}}%
 {\end{enumerate}}

\usepackage{listings}

\newcommand\JSONnumbervaluestyle{\color{blue}}
\newcommand\JSONstringvaluestyle{\color{red}}

\newif\ifcolonfoundonthisline

\makeatletter

\lstdefinestyle{json}
{
  showstringspaces    = false,
  keywords            = {false,true},
  alsoletter          = 0123456789.,
  morestring          = [s]{"}{"},
  stringstyle         = \ifcolonfoundonthisline\JSONstringvaluestyle\fi,
  MoreSelectCharTable =%
    \lst@DefSaveDef{`:}\colon@json{\processColon@json},
  basicstyle          = \ttfamily,
  keywordstyle        = \ttfamily\bfseries,
}

\newcommand\processColon@json{%
  \colon@json%
  \ifnum\lst@mode=\lst@Pmode%
    \global\colonfoundonthislinetrue%
  \fi
}

\lst@AddToHook{Output}{%
  \ifcolonfoundonthisline%
    \ifnum\lst@mode=\lst@Pmode%
      \def\lst@thestyle{\JSONnumbervaluestyle}%
    \fi
  \fi
  \lsthk@DetectKeywords%
}

\lst@AddToHook{EOL}%
  {\global\colonfoundonthislinefalse}

\makeatother
\usepackage{etoolbox}
\usepackage{natbib}
\usepackage{url}
\newcounter{bibcount}
\makeatletter
\patchcmd{\@lbibitem}{\item[}{\item[\hfil\stepcounter{bibcount}{[\thebibcount]}}{}{}
\renewcommand\NAT@bibsetup%
  [1]{\setlength{\leftmargin}{\bibhang}\setlength{\itemindent}{-\parindent}%
      \setlength{\itemsep}{\bibsep}\setlength{\parsep}{\z@}}
\makeatother

\definecolor{mybrown}{RGB}{128,64,0}

\definecolor{titlecolor}{HTML}{4c9cff}

\newcolumntype{L}{>{\raggedright\arraybackslash}p{0.74\linewidth}}
\begin{document}

\title{Rubrics to Tokens: Bridging Response-level Rubrics and Token-level Rewards in Instruction Following Tasks}

\author{
\textbf{Tianze Xu}$^{1,5}$ \quad
\textbf{Yanzhao Zheng}$^{2}$ \quad
\textbf{Pengrui Lu}$^{1,4,5}$ \quad
\textbf{Lyumanshan Ye}$^{1,5}$ \quad
\textbf{Yong Wu}$^{3}$ \\
\textbf{Zhentao Zhang}$^{2}$ \quad
\textbf{Yuanqiang Yu}$^{2}$ \quad
\textbf{Chao Ma}$^{2}$ \quad
\textbf{Jihuai Zhu}$^{2}$ \quad
\textbf{Pengfei Liu}$^{1,4,5}$\thanks{Corresponding author} \\
\textbf{Baohua Dong}$^{2}$ \quad
\textbf{Hangcheng Zhu}$^{2}$ \quad
\textbf{Ruohui Huang}$^{2}$ \quad
\textbf{Gang Yu}$^{2}$\\\;\\
\textsuperscript{1}Shanghai Jiao Tong University\quad
\textsuperscript{2}Alibaba Group\quad
\textsuperscript{3}Zhejiang University\\
\textsuperscript{4}Shanghai Innovation Institute\quad
\textsuperscript{5}GAIR\quad
}

\maketitle

\fancypagestyle{firstpage}{%
    \fancyhead{}
    \lhead{\lower0.70cm\hbox{\includegraphics[height=0.35cm]{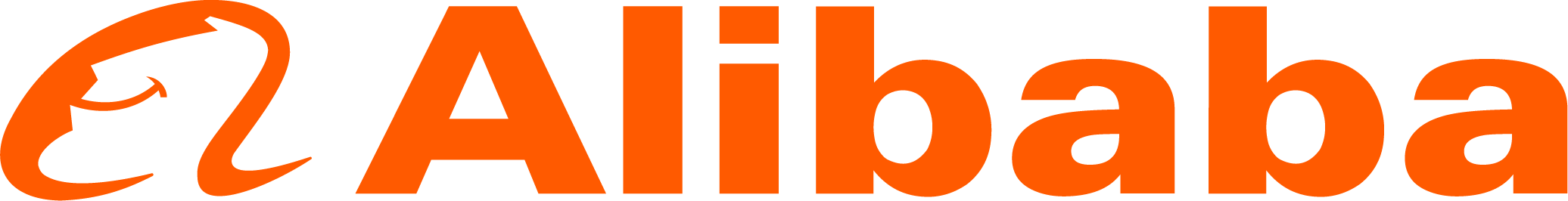}}}
    \rhead{\lower0.87cm\hbox{\includegraphics[height=0.7cm]{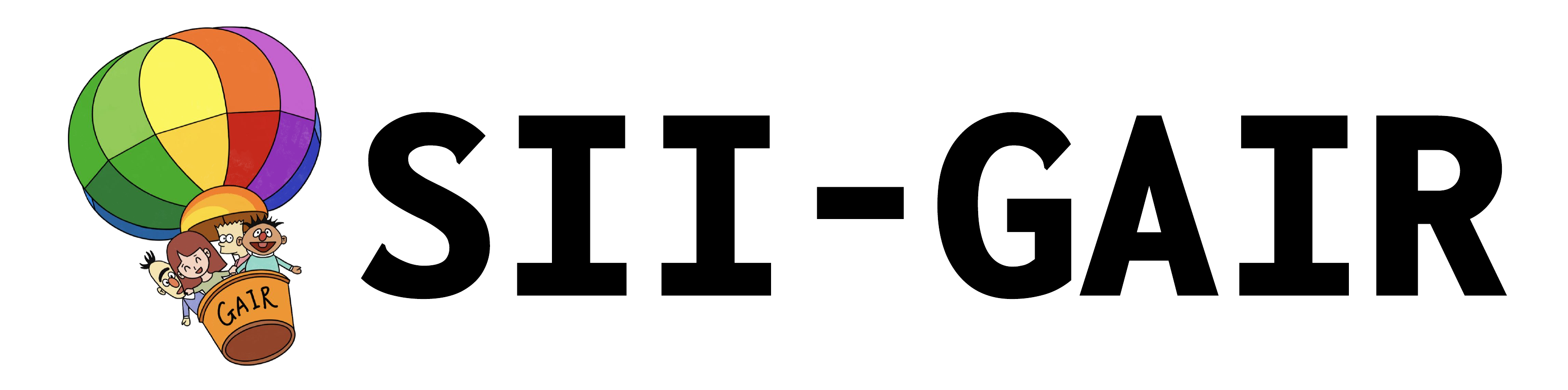}}}
    \renewcommand{\headrule}{\vskip 0.45cm \hrule width\headwidth height\headrulewidth}
    \renewcommand{\headrulewidth}{0pt}
}
\thispagestyle{firstpage}



\begin{abstract}
  Rubric-based Reinforcement Learning (RL) has emerged as a promising approach for aligning Large Language Models (LLMs) with complex, open-domain instruction following tasks. However, existing methods predominantly rely on response-level rewards, introducing severe reward sparsity and reward ambiguity problems. To address these issues, we propose \textbf{Rubrics to Tokens (RTT)}, a novel rubric-based RL framework that bridges coarse response-level scores and fine-grained token-level credit assignment. RTT introduces a \textbf{Token-Level Relevance Discriminator} to predict which tokens in the response are responsible for a specific constraint, and optimizes the policy model via \textbf{RTT-GRPO}, which integrates response-level and token-level advantages within a unified framework. Furthermore, we identify the \textbf{Group Partitioning Problem} when transitioning from one-dimensional, outcome-level reward to three-dimensional reward space in the token-level rubric-based RL, and propose a novel group normalization method, called \textbf{Intra-sample Token Group Normalization}, to accommodate this shift. Extensive experiments demonstrate that RTT consistently outperforms all baselines in both instruction-level and rubric-level accuracy across different models. We open-source RTT framework in \url{https://github.com/TURLEing/Rubrics-To-Tokens}.


\end{abstract} 

\pagestyle{fancy}
\lhead{\rightmark}
\rhead{}
\renewcommand{\headrulewidth}{0.7pt}
\setlength{\headsep}{5mm}



\section{Introduction}
\label{sec::intro}

In recent years, as Large Language Models (LLMs) tackle open-domain text generation tasks, such as instruction following~\citep{ouyang2022training, zhou2023ifeval, qin2024infobench}, Rubric-based Reinforcement Learning (Rubric-based RL) is emerging as a prominent direction~\citep{gunjal2025rubrics, he2025advancedif, goel2025training}. By defining structured and interpretable constraints, rubric-based rewards can capture multiple dimensions of quality beyond binary correctness, enabling effective alignment with nuanced open-domain objectives~\citep{huang2025reinforcement}. In this way, they provide a practical alternative to Reinforcement Learning from Verifiable Rewards (RLVR)~\citep{lambert2024tulu, guo2025deepseek} for unstructured, real-world reasoning tasks that lack easily verifiable answers.


However, the efficacy of current Rubric-based RL paradigms is heavily bottlenecked by the response-level reward design strategies~\citep{zhang2025replay}. Existing approaches typically rely on either strict binary rewards~\citep{gunjal2025rubrics, he2025advancedif}, where a response is rewarded only if it perfectly meets every constraint, or aggregated rewards based on the ratio of satisfied constraints~\citep{huang2025reinforcement, goel2025training}. The binary approach suffers from reward sparsity problem; because RL relies on self-exploration, an initial model with limited capabilities  struggles to generate responses that satisfy all constraints, leaving it without meaningful gradient signals~\citep{zhou2025breaking}. Conversely, while aggregated rewards can stabilize early training, they introduce reward ambiguity problems; two vastly different responses can receive identical scores for satisfying disjoint subsets of constraints, producing coarse credit assignment that confounds training~\citep{zhang2025replay}.

To resolve these issues, we propose Rubrics to Tokens (RTT), a novel Rubric-based RL training framework that bridges response-level rubric scores and token-level credit assignment. This method is grounded in a straightforward hypothesis: \textbf{For many specific constraints, only a subset of tokens plays a critical role in their satisfaction or violation}. For example, given a constraint that dictates ``Do not provide specific mathematical formulas," the penalty should exclusively target the tokens constituting the formula, without affecting the reward signals for the rest of the well-formed response. By introducing a Rubric-to-Token aligned reward signal, RTT enables highly precise, fine-grained advantage computation. 

\begin{figure}[t]
    \centering
    \includegraphics[width=1\linewidth]{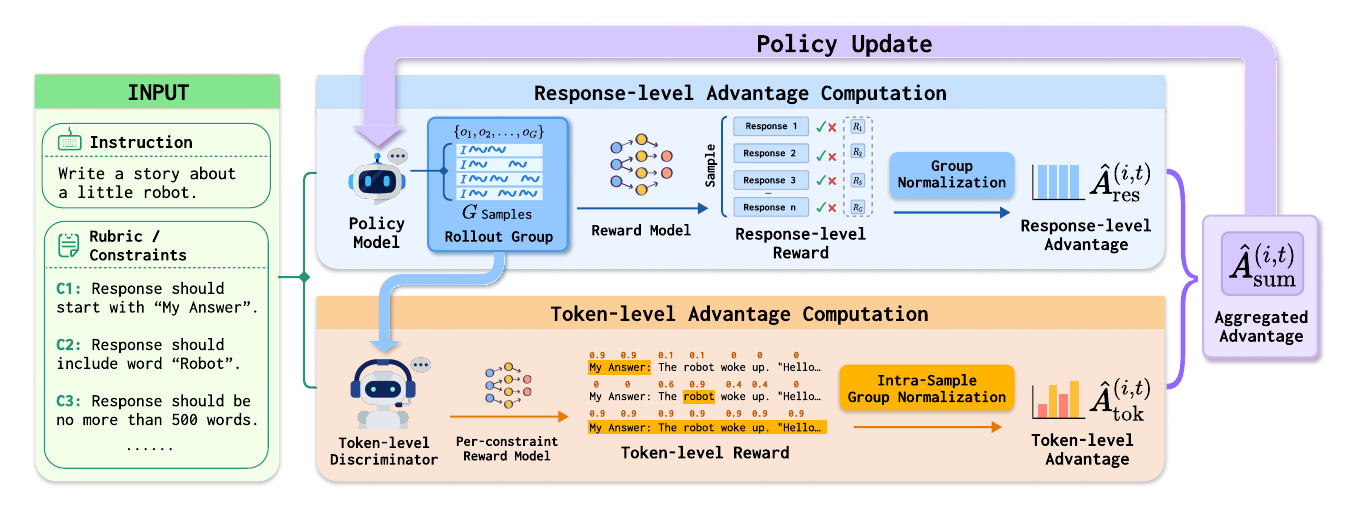}
    \caption{Architecture of the RTT framework. The policy model first generates a response group for instructions. The top computes $\smash{\hat{A}_{res}^{(i,t)}}$ by normalizing rewards across the sampled responses. The bottom uses a token-level discriminator to map rewards to specific tokens based on rubrics, applying Intra-Sample Group Normalization to calculate $\smash{\hat{A}_{tok}^{(i,t)}}$. These two components are merged into the final aggregated advantage $\smash{\hat{A}_{sum}^{(i,t)}}$ to drive the policy update.}
    \label{fig:overview}
\end{figure}

Specifically, we first train a Token-Level Relevance Discriminator that predicts which tokens in a response are responsible for the satisfaction or violation of each constraint,  producing token-level reward signals. Next, we adapt GRPO~\citep{shao2024deepseekmath} into RTT-GRPO, a training objective that jointly optimizes over response-level and token-level advantages, as shown in Figure~\ref{fig:overview}. We further identify a normalization design challenge in the three-dimensional reward space under token-level rubric-based RL scenario, namely how to define normalization boundaries across different reward dimensions, which we term the \textbf{Group Partitioning Problem}; to address it, we propose \textbf{Intra-sample Token Group Normalization} as the solution. Extensive experiments across multiple models and benchmarks demonstrate that RTT consistently outperforms all baselines in both instruction-level and rubric-level accuracy. In summary, our core contributions are as follows:
\begin{itemize}[itemsep=0.5pt, topsep=1pt]
    \item We propose \textbf{Rubrics to Tokens (RTT)}, a Rubric-based RL framework that utilizes a Token-Level Relevance Discriminator to produce token-level reward signal, and introduces RTT-GRPO to jointly optimize response-level and token-level advantages within a unified framework.
    \item We identify the \textbf{Group Partitioning Problem} in three-dimensional reward space of token-level rubric-based RL about how to define normalization boundaries across different dimensions, and propose \textbf{Intra-sample Token Group Normalization} to adress it.
    \item Extensive experiments demonstrate that RTT consistently outperforms all baselines across instruction-following benchmarks, while preserving general capacities. Furthermore, we open-source our RTT framework.
\end{itemize}

\section{Related Work}
\label{sec::rel_work}
\subsection{Instruction Following Methods}

LLM instruction following has progressed from Supervised Fine-Tuning (SFT), which learns desired behaviors from high-quality data~\citep{zhou2023lima, lou2023muffin, xu2023wizardlm}, to alignment methods such as DPO~\citep{rafailov2023direct} and Reinforcement Learning from Human Feedback (RLHF)~\citep{ouyang2022training, cheng2024spar, dong2024self}. To reduce reliance on human annotation, recent work~\citep{peng2025verif, liu2025recast} explores Reinforcement Learning from AI Feedback (RLAIF) through the LLM-as-a-Judge paradigm~\citep{zheng2023judging, srivastava2025technical}.

Despite these advancements, complex, multi-constraint instruction following remains a critical bottleneck~\citep{he2025advancedif}, as initial models often fail to satisfy all constraints~\citep{he2024complex, wen2024benchmarking}. Recent works~\citep{zhang2025replay} explore hindsight relabeling to extract meaningful learning signals from failed attempts.

\subsection{Token-level Credit Assignment}

In Reinforcement Learning (RL) for LLMs, outcome reward is commonly used as the reward signal, providing coarse supervision in complex, long-horizon tasks~\citep{zeng2025reinforcing, ping2026longr}. On the other side, dense process reward, especially token-level reward, has been proved to improve credit assignment and training efficiency~\citep{lightman2023let, wang2024math, cui2025process}. Existing token-level methods mainly either use external discriminators to produce span-level annotations for token credit assignment~\citep{yoon2024tlcr, guo2023beyond, cao2024beyond}, or incorporate auxiliary signals such as policy entropy and implicit rewards to reweight tokens during training~\citep{cui2025process, tan2025gtpo}.  

\subsection{Rubric-based Reinforcement Learning}
To extend RLVR beyond domains with strictly verifiable outcomes like mathematics~\citep{shao2024deepseekmath} or code generation~\citep{jiang2025coderl}, the community is generally adopting Rubric-based Reinforcement Learning, by using structured rubrics and LLM judges to provide reward signals for open-ended tasks~\citep{gunjal2025rubrics, huang2025reinforcement}. Prior work applies instance-specific rubrics to open-ended RLVR~\citep{gunjal2025rubrics}, constructing large-scale rubric systems~\citep{huang2025reinforcement}, developing rubric-driven benchmarks~\citep{sharma2025researchrubrics, he2025advancedif}, and applying rubric-guided self-grading to scientific planning tasks~\citep{goel2025training}. However, existing approaches predominantly aggregate rubric satisfaction into a single, scalar response-level reward, which introduces reward sparsity and ambiguity problems~\citep{zhou2025breaking, zhang2025replay}.

\section{Preliminaries and Backgrounds}
\label{sec::preliminaries}

\subsection{Instruction Following}
\label{sec::preliminary_if}

Let $q$ denote an instruction and $y\sim \pi_\theta( \cdot \mid q)$ be a sample response from a model parameterized by $\theta$. Each instruction consists of a task $x$ and a rubric, defined as a set of constraints $\mathcal{C} = \{c_1, c_2, \dots, c_n\}$. Following \cite{peng2025verif}, we divide $\mathcal{C}$ into \textbf{hard constraints}, verifiable by rules (e.g., exact length limits or format requirements), and \textbf{soft constraints}, which require nuanced semantic evaluation (e.g., specific writing styles or tones).

We adopt a hybrid evaluation approach to verify the constraint satisfaction~\citep{peng2025verif, zhang2025replay}: hard constraints use deterministic rule-based verifiers, while soft constraints are evaluated via LLM-as-a-Judge. Formally, we denote this evaluation by the indicator function $\mathbb{I}(q, y, c_i)$:
\begin{equation}
\mathbb{I}(q, y, c_i) =
\begin{cases}
    \text{Rule}(c_i, y), & \text{if } c_i \in \mathcal{C}_{\text{hard}}, \\
    \text{LLM}(c_i, y), & \text{if } c_i \in \mathcal{C}_{\text{soft}},
\end{cases}
\end{equation}

where $C_\text{hard}$ and $C_\text{soft}$ denote the sets of hard and soft constraints respectively. The specific evaluation prompt utilized for the judge LLM is detailed in Appendix~\ref{app:judge_prompt}.

While extending individual constraint to the entire set of constraints, there are two aggregation strategies that measure performance from different granularities of instruction following tasks~\citep{he2025advancedif}: All-or-Nothing and Constraint Satisfaction Rate.

\paragraph{All-or-Nothing Metric (AON).} This metric enforces a strict standard where a response is correct only if it satisfies \textit{every} constraint associated with the instruction $q$. It is defined as the logical conjunction of all individual evaluations:

\begin{equation}
    \text{Score}(q, y) = \prod_{i=1}^{|\mathcal{C}|} \mathbb{I}(q, y, c_i) =
    \begin{cases}
        1, & \text{if } \forall c_i \in \mathcal{C}, \mathbb{I}(q, y, c_i) = 1, \\
        0, & \text{otherwise}.
    \end{cases}
\end{equation}

\paragraph{Constraint Satisfaction Rate (CSR).} This metric measures the model's ability to follow individual constraints, which is calculated as the percentage of satisfied constraints:
\begin{equation}
    \text{Score}(q, y) = \frac{1}{|\mathcal{C}|} \sum_{i=1}^{|\mathcal{C}|} \mathbb{I}(q, y, c_i).
\end{equation}

These two metrics provide a complementary evaluation framework, however, face their inherent challenges when employed as reward signals in the reinforcement learning. AON metric suffers from reward sparsity problems, while CSR metric introduces a severe risk of reward ambiguity~\citep{peng2025verif, zhang2025replay}.


\begin{figure}[t]
    \centering
    \includegraphics[width=\linewidth]{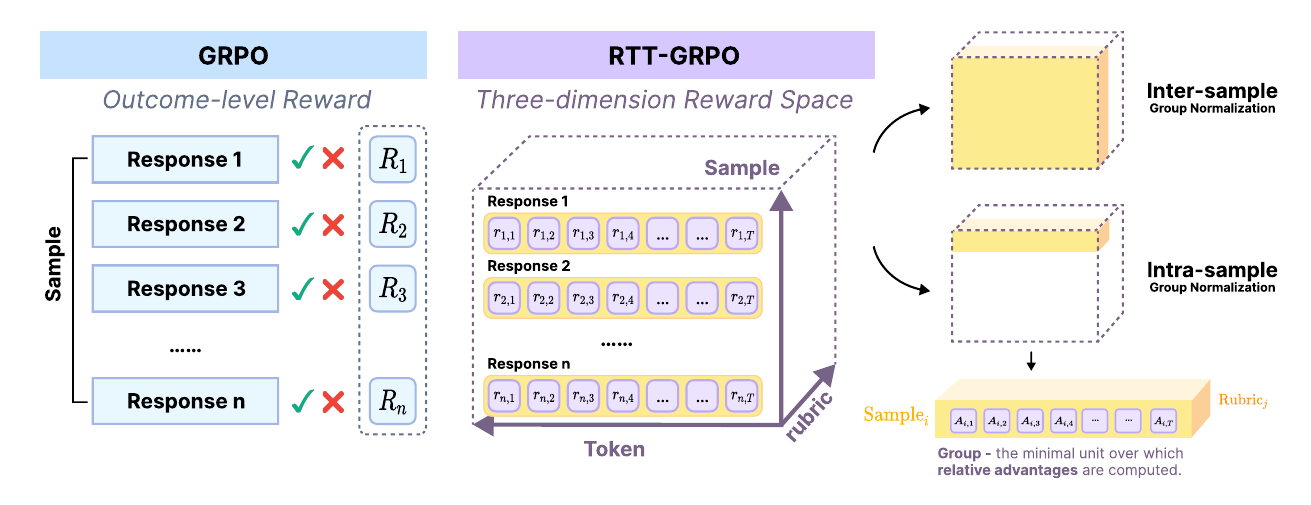}
    \caption{Reward structure under standard GRPO and RTT-GRPO. In GRPO (left), each of $G$ sampled completions receives one outcome-level reward, advantages are normalized along the \textit{sample} axis only. The RTT-GRPO (right) assigns rewards over three dimensions, motivating two group normalization strategies, \emph{inter-sample} and \emph{intra-sample} group normalization.}
    \label{fig:reward_structure}
\end{figure}

\subsection{GRPO and Group Partitioning Problem}
\label{sec::grpo_group}

\textbf{Group Relative Policy Optimization (GRPO)}~\citep{shao2024deepseekmath} computes policy gradients through group-relative advantage normalization, eliminating the need for a critic model. In this paper, we define a \textbf{group} as the minimal unit for computing relative advantages. In standard GRPO, a group contains $G$ responses $\{o_1, \dots, o_G\}$ sampled from the same instruction $q$. The response-level reward $r_i = \text{Score}(q, o_i)$ is normalized within the group to yield a scalar advantage $\smash{\tilde{r}_i = \frac{r_i - \text{mean}(r)}{\text{std}(r)}}$, which is uniformly applied to all tokens in $o_i$. The group space is therefore one-dimensional, varying only along the sample axis.

In Token-level Rubric-Based RL, however, the reward signal spans three axes: \textit{sample} (which response), \textit{token} (which position), and \textit{criteria} (which constraint). As shown in Figure~\ref{fig:reward_structure}, the group space is thus expanded from one-dimensional, outcome-level axis into three-dimensional space, which raises a question: how to define group boundaries in multi-dimensions for relative advantages to yield meaningful relative advantages.

We formalize this as the \textbf{Group Partitioning Problem}. While prior work studied group size and intra-group sampling~\citep{zhou2026demystifying, shrivastava2025sample, gu2025group}, we are the first to study group partitioning in high-dimensional reward spaces.

A natural strategy is \textbf{Inter-Sample Token Group Normalization}: for each constraint $c_k$, compute token-level scores across all $G$ responses, normalize them jointly over all tokens of all samples to obtain per-criteria advantages, and average them into the final token-level advantage. However, this strategy suffers from length bias problem: when response lengths differ within a group, the joint normalization is \textbf{dominated by longer responses} as they contribute more tokens, inflating the mean and variance estimates and suppresses signals from shorter responses. Formal definition and length bias problem are discussed in Appendix~\ref{app:inter_sample_norm}.

To address this, we propose \textbf{Intra-sample Token Group Normalization}, which normalizes within individual response rather than jointly. This group normalization strategy eliminates length-induced bias and decouples token-level advantages from response-level advantages. The detailed formulation is provided in Section~\ref{sec::token_advantage}.

\section{Rubrics To Tokens}
\label{sec::rtt}

In this section, we introduce the \textbf{Rubrics to Tokens (RTT)} framework. RTT leverages a Token-Level Relevance Discriminator to project coarse response-level rubric scores into fine-grained token-level rewards, and optimizes the policy via RTT-GRPO with intra-sample token group normalization.

\subsection{Token-Level Relevance Discriminator}

A fundamental limitation of response-level rubric reward is the inability to assign credit to the specific tokens responsible for the generated response. To align response-level rewards with token-level behaviors, we introduce a Token-Level Relevance Discriminator. 

Formally, given a prompt $x$ (detailed in Appendix~\ref{app:infer_prompt}), a constraint $c_i \in \mathcal{C}$, and a generated response $y = (y_1, y_2, \dots, y_T)$ of $T$ tokens, the discriminator $D_\phi$ takes the concatenation $(x, c_i, y)$ as input and produces a token-level relevance score for each position:
\begin{equation}
    D_\phi(x, c_i, y) = \{ p_1, p_2, \dots, p_T \} \in [0, 1]^T,
\end{equation}
where $p_t$ denotes the predicted relevance score of token $y_t$ to constraint $c_i$.

\paragraph{Data Collection and Negative Sampling.} To train this discriminator, we construct a high-quality token-level dataset. We firstly collect diverse prompts and corresponding constraints from established instruction-following datasets, including Tulu-3~\citep{lambert2024tulu} and HiR-16K~\citep{zhang2025replay}. Detailed data sources and construction are provided in Appendix~\ref{app:data_sources}.

A robust discriminator must identify tokens relevant to constraint $c_i$, whether the response satisfies or violates it. Thus, constructing high-quality negative samples (i.e., responses that fail the constraint) is essential. We employ two strategies to generate negative samples:
\begin{itemize}[itemsep=0.5pt, topsep=1pt]
    \item \textbf{Minimal Modification:} we instruct an LLM to rewrite a compliant response to violate the constraint with minimal modification, forcing the discriminator to learn fine-grained distinctions.
    \item \textbf{Constraint Omission:} during generation, we intentionally withhold a specific constraint; the resulting response probably fails the unseen constraint, providing organic and natural negative examples.
\end{itemize}

\paragraph{Annotation Strategy.} After the response generation, we use a strong LLM to automatically annotate relevant text segments in response  $y$ for constraint $c_i$ following a taxonomy strategy, and map them into corresponding tokens, detailed in Appendix~\ref{app:annotation_taxonomy}.

\paragraph{Model Architecture and Training Objective.} The discriminator $D_\phi$, takes the concatenated prompt with constraint, and response as input. A token-level classification head applied to the final hidden state at position $t$ outputs an unnormalized logit $\hat{s}_t$, from which the predicted relevance probability is obtained via sigmoid: $p_t = \sigma(\hat{s}_t) \in [0, 1]$. We optimize this token-wise binary classification task via Binary Cross-Entropy (BCE) loss:
\begin{equation}
\mathcal{L}_{\text{BCE}}(\phi) = - \frac{1}{|V|} \sum_{t \in V} \left[ l_t \log(p_t) + (1 - l_t) \log(1 - p_t) \right],
\end{equation}

where $V$ is the set of valid token indices (excluding prompt tokens and padding), and $l_t \in \{0, 1\}$ is the ground-truth relevance label derived from our LLM annotator.

\subsection{Rubric-to-Token RL Training}
\label{sec::rtt_rl_training}

\subsubsection{Rubric-to-Token GRPO}
\label{sec::rtt_grpo}

To integrate Rubric to Token into LLM reinforcement learning, we adapt Group Relative Policy Optimization (GRPO) algorithm~\citep{shao2024deepseekmath} into \textbf{Rubric-to-Token GRPO (RTT-GRPO)}, an objective function compatible with dense, token-level reward signals.

Specifically, for each instruction $q$, RTT-GRPO samples a group of outputs $\{o_1, o_2, \dots , o_G \}$ from the old policy $\pi_{\theta_{\text{old}}}$ , and a reward model is used to score each output, yielding $G$ rewards $\smash{r = \{r_1, r_2, \dots , r_G \}}$, consistent with original GRPO in \cite{shao2024deepseekmath}. The policy model is then optimized by maximizing the following objective:
\begin{equation}
\begin{aligned}
\mathcal{J}_{\text{RTT-GRPO}}(\theta) & = \mathbb{E} [q \sim P(q),~\{o_i\}_{i = 1}^G\sim\pi_{\theta_{old}}]  \\ &\quad \frac{1}{\sum_{i = 1}^G |o_i|} \sum_{i = 1}^G \sum_{t = 1}^{|o_i|} \min \left(w_{i,t}(\theta) \hat{A}^{(i,t)}_{\text{sum}},~ \text{clip}(w_{i,t}(\theta), 1-\epsilon, 1+\epsilon) \hat{A}^{(i,t)}_{\text{sum}}\right) ,
\end{aligned}    
\end{equation}

where $w_{i,t}(\theta) = \frac{\pi_\theta(o_{i,t}|q, o_{i,<t})}{\pi_{\theta_{\text{old}}}(o_{i,t}|q, o_{i,<t})}$ is the standard importance sampling weight. The combined advantage for the $t$-th token in the $i$-th response, denoted as $\smash{\hat{A}^{(i,t)}_{\text{sum}}}$, integrates both response-level and token-level advantages:
\begin{equation}
\hat{A}^{(i,t)}_{\text{sum}} = \alpha \hat{A}^{(i,t)}_{\text{res}} + \beta \smash{\hat{A}^{(i,t)}_{\text{tok}}} ,
\label{eq:combined_advantage}
\end{equation}

where $\alpha$ and $\beta$ are hyper-parameters controlling the contribution of reward components. 

The response-level advantage $\smash{\hat{A}^{(i,t)}_{\text{res}}}$ is calculated as $\smash{\hat{A}^{(i,t)}_{\text{res}}=\tilde{r}_i = \frac{r_i - \text{mean}(r)}{\text{std}(r)}}$, where $r_i=\text{Score}(q, o_i)$, providing comparative signals across different rollout responses.  Depending on different scoring granularities, we employ either AON or CSR metrics to calculate response-level rewards. 

The token-level advantage $\hat{A}^{(i,t)}_{\text{tok}}$ dynamically maps the constraint satisfaction signals back to the trajectory, via Intra-sample Token Group Normalization, to redistribute credit within a single response. Section \ref{sec::token_advantage} below details formulation and derivation of $\smash{\hat{A}^{(i,t)}_{\text{tok}}}$.

\subsubsection{Intra-sample Token Group Normalization}
\label{sec::token_advantage}

Given a constraint $c_k \in \mathcal{C}$ and a generated response $o_i$ of length $T$, we first employ the token-level discriminator to predict the relevance of each token. This yields a sequence of relevance probabilities for the entire response, denoted as $\smash{P = \{p^{(i,t)}_{c_k} \mid t=1, \dots, T\}}$. To project the binary response-level $\text{Score}(o_i, c_k)$ onto individual tokens, we multiply $\text{Score}(o_i, c_k)$ by the relevance sequence $\text{P}$, producing token-level rewards $\smash{R_{c_k}^{(i)} = \{r^{(i,t)}_{c_k} \mid t=1, \dots, T\}}$ for the specific constraint, where $\smash{r^{(i,t)}_{c_k} = \text{Score}(o_i, c_k) \cdot p^{(i,t)}_{c_k}}$, and $\smash{\text{Score}(o_i, c_k)=2\cdot\bigl(\mathbb{I}(q, o_i, c_k)-0.5\bigr)}$. 

To compute the token-level advantage while decoupling it from the response-level reward, we apply a localized normalization strategy. \textbf{Specifically, we define the normalization group as the set of all $T$ tokens within a single response $o_i$.} The advantage for a single constraint is computed by standardizing the rewards: $\smash{A_{c_k}^{(i,t)} = (r^{(i,t)}_{c_k} - \mu_R) / \sigma_R}$, where $\mu_R$ and $\sigma_R$ are the mean and standard deviation of $\smash{R_{c_k}^{(i)}}$. This intra-response group normalization adapts to token distributions by amplifying  advantage for relevant tokens, while naturally neutralizing for those uniformly relevant or irrelevant tokens in the responses.

Finally, for an instruction with multiple constraints, we aggregate the signals by averaging the normalized advantages across the entire constraint set $\mathcal{C}$, yielding the final token-level advantage $\smash{\hat{A}_\text{tok}^{(i,t)} = \frac{1}{|\mathcal{C}|} \sum_{c_k \in \mathcal{C}} A_{c_k}^{(i,t)}}$.

\section{Experimental Settings}
\label{sec::experiment_setting}

\subsection{Datasets}
\label{sec::datasets}

We use HiR-16K~\citep{zhang2025replay} as our training dataset, which aggregates multiple sources to ensure broad instruction-following coverage. Specifically, HiR-16K includes MulDimIF~\citep{ye2025multi} for diverse constraint categories, VerIF~\citep{peng2025verif} for constraint-verifiable examples, IFTrain~\citep{pyatkin2025ifbench} from the IFBench benchmark, and Chatbot Arena~\citep{zheng2023judging} for real-world human queries.

\vspace{-5pt}
\subsection{Benchmarks}
\label{sec::benchmarks}

We evaluate RTT on two categories of benchmarks. For in-domain instruction-following, we report results on IFEval~\citep{zhou2023ifeval}, IFBench~\citep{pyatkin2025ifbench},  MulDimIF~\citep{ye2025multi}, and AdvancedIF~\citep{he2025advancedif}. To assess generalization beyond the instruction-following domain, we additionally evaluate on three out-of-domain benchmarks: MATH-500~\citep{lightman2023let}, GPQA~\citep{rein2024gpqa}, and MMLU-Pro~\citep{wang2024mmluPro}. Detailed benchmark configurations, including evaluation protocols and prompt formats, are provided in Appendix~\ref{app:benchmarks}.

\subsection{Models and Configurations}
\label{sec::models}

We train two Token-Level Relevance Discriminators, initialized from Qwen3-4B-Base and Llama3.2-3B-Base, to assess generality across model families. Policy models include Qwen3-4B-Instruct, Qwen2.5-7B-Instruct~\citep{qwen2025qwen25technicalreport, yang2025qwen3}, and Llama3.2-3B-Instruct. We report results for both RTT-CSR and RTT-AON variants for each policy model, using CSR and AON metrics as the response-level reward. For the implementation of RTT-GRPO, we set $\alpha=1$, $\beta=0.5$ in Equation~(\ref{eq:combined_advantage}). All reinforcement learning experiments are conducted within the ROLL framework~\citep{wang2025reinforcement}. For soft constraints ($\mathcal{C}_{\text{soft}}$), we employ DeepSeek-V3~\citep{liu2024deepseek} as the judge LLM. Full hyperparameters and prompts are provided in Appendix~\ref{app:hyperparams} and \ref{app:judge_prompt}.

\subsection{Baselines and Evaluation Metrics}
\label{sec::baselines}

We compare RTT against three baselines: 1) Supervised Fine-Tuning (SFT) on Gemini-3-Pro generated responses of the training data; 2) Direct Preference Optimization (DPO)~\citep{rafailov2023direct} on pairs generated by Gemini-3-Pro (chosen responses) and Qwen2.5-7B-Instruct (rejected responses); and 3) two RL baselines using All-or-Nothing (RL-AON) and Constraint Satisfaction Rate (RL-CSR) metric as the response-level rewards.

For each benchmark except MulDimIF, we report results in two metrics: \textbf{Instruction-level Accuracy} measures the fraction of test prompts that satisfies all constraints, reflecting holistic performance for multi-constraint instructions; while \textbf{Rubric-level Accuracy} measures the average constraint satisfaction rate, providing a fine-grained view of per-constraint performance. Detailed metrics for each benchmark are provided in Appendix~\ref{app:benchmarks}.


\section{Results}
\label{sec::result}

\begin{table}[!t]
    \centering
    \resizebox{\textwidth}{!}{%
    \begin{tabular}{llllllll}
    \toprule
    & \multicolumn{2}{c}{\textbf{IFEval}} & \multicolumn{2}{c}{\textbf{IFBench}} & \multicolumn{1}{c}{\textbf{MulDimIF}} & \multicolumn{2}{c}{\textbf{AdvancedIF}} \\
    \cmidrule(lr){2-3}\cmidrule(lr){4-5}\cmidrule(lr){6-6}\cmidrule(lr){7-8}
    \textbf{Model} & \multicolumn{1}{c}{Prompt} & \multicolumn{1}{c}{Instruct} & \multicolumn{1}{c}{Prompt} & \multicolumn{1}{c}{Instruct} & \multicolumn{1}{c}{Acc.} & \multicolumn{1}{c}{Overall} & \multicolumn{1}{c}{Rubric} \\
    \midrule
    Qwen3-4B-Instruct$^\dagger$        & 83.40 \nc{0.0} & 88.13 \nc{0.0} & 30.95 \nc{0.0} & 32.83 \nc{0.0} & 56.08 \nc{0.0} & 44.31 \nc{0.0} & 79.17 \nc{0.0} \\
    ~~+ SFT                  & 83.73 \up{0.3} & 88.85 \up{0.7} & 29.59 \dn{1.4} & 31.64 \dn{1.2} & 56.50 \up{0.4} & 45.07 \up{0.8} & 79.23 \up{0.1} \\
    ~~+ DPO                  & 82.62 \dn{0.8} & 87.77 \dn{0.4} & 29.93 \dn{1.0} & 32.84 \nc{0.0} & 55.83 \dn{0.3} & 45.71 \up{1.4} & 79.33 \up{0.2} \\
    ~~+ RL-CSR               & 84.29 \up{0.9} & \underline{89.21} \up{1.1} & 32.31 \up{1.4} & 33.43 \up{0.6} & 74.38 \up{18.3} & 46.02 \up{1.7} & 79.81 \up{0.6} \\
    ~~+ RL-AON               & 82.99 \dn{0.4} & 88.61 \up{0.5} & 32.99 \up{2.0} & 35.52 \up{2.7} & 74.25 \up{18.2} & 47.29 \up{3.0} & 79.69 \up{0.5} \\
    \rowcolor{rowblue}
    ~~\textbf{+ RTT-CSR (Ours)} & \underline{85.03} \up{1.6} & \textbf{90.17} \up{2.0} & \underline{34.01} \up{3.1} & \underline{36.12} \up{3.3} & \underline{76.33} \up{20.3} & \textbf{48.39} \up{4.1} & \underline{80.77} \up{1.6} \\
    \rowcolor{rowblue}
    ~~\textbf{+ RTT-AON (Ours)} & \textbf{85.21} \up{1.8} & 88.73 \up{0.6} & \textbf{34.69} \up{3.7} & \textbf{37.61} \up{4.8} & \textbf{76.75} \up{20.7} & \underline{47.48} \up{3.2} & \textbf{80.79} \up{1.6} \\
    \midrule
    Qwen2.5-7B-Instruct$^\dagger$      & 69.68 \nc{0.0} & 77.93 \nc{0.0} & 25.85 \nc{0.0} & 27.16 \nc{0.0} & 51.08 \nc{0.0} & 29.48 \nc{0.0} & 71.31 \nc{0.0} \\
    ~~+ SFT                  & 71.90 \up{2.2} & 79.38 \up{1.5} & 28.23 \up{2.4} & 30.15 \up{3.0} & 52.83 \up{1.8} & 29.71 \up{0.2} & 69.61 \dn{1.7} \\
    ~~+ DPO                  & 71.16 \up{1.5} & 78.90 \up{1.0} & 27.55 \up{1.7} & 28.96 \up{1.8} & 51.08 \nc{0.0} & 30.82 \up{1.3} & 70.90 \dn{0.4} \\
    ~~+ RL-CSR               & 78.56 \up{8.9} & 84.41 \up{6.5} & 28.23 \up{2.4} & 29.85 \up{2.7} & 66.67 \up{15.6} & 30.91 \up{1.4} & 72.00 \up{0.7} \\
    ~~+ RL-AON               & 79.67 \up{10.0} & 85.13 \up{7.2} & \underline{31.97} \up{6.1} & 33.13 \up{6.0} & 69.41 \up{18.3} & 32.10 \up{2.6} & 72.33 \up{1.0} \\
    \rowcolor{rowblue}
    ~~\textbf{+ RTT-CSR (Ours)} & \underline{81.15} \up{11.5} & \underline{86.33} \up{8.4} & \textbf{32.31} \up{6.5} & \textbf{34.63} \up{7.5} & \underline{69.67} \up{18.6} & \textbf{33.74} \up{4.3} & \textbf{73.72} \up{2.4} \\
    \rowcolor{rowblue}
    ~~\textbf{+ RTT-AON (Ours)} & \textbf{82.26} \up{12.6} & \textbf{87.29} \up{9.4} & \textbf{32.31} \up{6.5} & \underline{33.43} \up{6.3} & \textbf{71.83} \up{20.6} & \underline{32.51} \up{3.0} & \underline{72.70} \up{1.4} \\
    \midrule
    Llama3.2-3B-Instruct$^\star$     & 70.61 \nc{0.0} & 78.30 \nc{0.0} & 21.77 \nc{0.0} & 24.18 \nc{0.0} & 35.42 \nc{0.0} & 33.37 \nc{0.0} & 72.01 \nc{0.0} \\
    ~~+ SFT                  & 69.50 \dn{1.1} & 77.82 \dn{0.5} & 22.79 \up{1.0} & 24.78 \up{0.6} & 36.50 \up{1.1} & 32.89 \dn{0.5} & 72.18 \up{0.2} \\
    ~~+ DPO                  & 70.61 \nc{0.0} & 78.78 \up{0.5} & 24.15 \up{2.4} & 26.27 \up{2.1} & 34.50 \dn{0.9} & 32.28 \dn{1.1} & 71.14 \dn{0.9} \\
    ~~+ RL-CSR               & 68.21 \dn{2.4} & 77.82 \dn{0.5} & 23.81 \up{2.0} & 25.97 \up{1.8} & 56.92 \up{21.5} & 36.29 \up{2.9} & 73.47 \up{1.5} \\
    ~~+ RL-AON               & \underline{76.16} \up{5.6} & \underline{83.33} \up{5.0} & 24.15 \up{2.4} & 25.37 \up{1.2} & \underline{66.83} \up{31.4} & 35.32 \up{2.0} & 73.53 \up{1.5} \\
    \rowcolor{rowblue}
    ~~\textbf{+ RTT-CSR (Ours)} & 74.49 \up{3.9} & 81.77 \up{3.5} & \underline{25.51} \up{3.7} & \underline{26.87} \up{2.7} & 66.33 \up{30.9} & \underline{37.57} \up{4.2} & \textbf{74.37} \up{2.4} \\
    \rowcolor{rowblue}
    ~~\textbf{+ RTT-AON (Ours)} & \textbf{79.48} \up{8.9} & \textbf{85.37} \up{7.1} & \textbf{26.19} \up{4.4} & \textbf{27.46} \up{3.3} & \textbf{68.58} \up{33.2} & \textbf{37.99} \up{4.6} & \underline{74.07} \up{2.1} \\
    \bottomrule
    \end{tabular}%
    }
    \caption{Main results on instruction-following benchmarks. Fine-tuned method rows show scores with delta (\up{}/\dn{}) relative to the base model. Highlighted rows are our proposed RTT variants. Best results per model group are \textbf{bolded} and second-best are \underline{underlined}. $^\dagger$ We use discriminator initialized from Qwen3-4B-Base, $^\star$ initialized from Llama3.2-3B-Base.}
    \label{tab:main_results}
\end{table}

\subsection{Main Results}
\label{sec::main_results}

\textbf{RTT achieves the best instruction-level and rubric-level accuracy across all benchmarks.} Table~\ref{tab:main_results} presents the main results across all in-domain instruction-following benchmarks. Specifically, RTT-CSR and RTT-AON attain the highest scores on both instruction-level and rubric-level metrics across all benchmarks. Compared to the strongest RL baselines (RL-AON and RL-CSR), RTT yields average gains of 2.50\% in instruction-level accuracy and 1.64\% in rubric-level accuracy. Moreover, RTT improves both AON and CSR scores by 1.58\% and 2.17\% respectively, further demonstrating the generality of the method. These improvements suggest that token-level credit assignment helps resolve the sparsity and ambiguity of response-level rewards. 


\begin{wraptable}{r}{0.57\textwidth}
\centering
\vspace{-10pt}
\small
\begin{tabular}{l|lll}
\toprule
\textbf{Model} & \textbf{Math-500} & \textbf{GPQA} & \textbf{MMLU-Pro} \\
\midrule
Qwen3-4B-Instruct        & 87.20 & 51.79 & 69.60 \\
\rowcolor{rowblue}
~~\textbf{+ RTT-AON (Ours)}     & \textbf{91.80} \up{4.6} & \textbf{52.01} \up{0.2} & 66.76 \dn{2.8} \\
\midrule
Qwen2.5-7B-Instruct      & 74.80 & 30.80 & 56.43 \\
\rowcolor{rowblue}
~~\textbf{+ RTT-AON (Ours)}     & 73.80 \dn{1.0} & \textbf{31.25} \up{0.5} & 56.12 \dn{0.3} \\
\midrule
Llama3.2-3B-Instruct     & 36.60 & 25.89 & 38.96 \\
\rowcolor{rowblue}
~~\textbf{+ RTT-AON (Ours)}     & \textbf{40.00} \up{3.4} & \textbf{26.12} \up{0.2} & \textbf{39.04} \up{0.1} \\
\bottomrule
\end{tabular}
\caption{RTT on out-of-domain benchmarks.}
\label{tab:ood_results}
\vspace{-10pt}
\end{wraptable}

\textbf{RTT generally preserves out-of-domain ability.} As shown in Table~\ref{tab:ood_results}, although RTT is trained solely on instruction-following data, it largely preserves or improves the models' OOD performance. These results suggest that RTT's fine-grained credit assignment regularizes the policy toward better intent grounding and constraint satisfaction without collapsing general reasoning ability.

\subsection{Ablation Study}
\label{sec::ablation}

\subsubsection{A1: Group Partitioning Strategies}
\label{sec::ablation_group}

We conduct a systematic ablation over the group partitioning strategies described in Section~\ref{sec::grpo_group}, comparing two representative strategies across three model configurations: (1) Inter-sample Token Group Normalization, where advantage is computed jointly across all tokens of all responses in a group; and (2) Intra-sample Token Group Normalization, which performs per-response normalization independently before aggregating across criteria. As Table~\ref{tab:ablation_group} indicates, Intra-sample Token Group Normalization strategy generally achieves the best performance across all models, confirming that per-response normalization eliminates length-induced bias in advantage estimation.

\begin{table}[!htbp]
\centering
\small
\resizebox{\textwidth}{!}{%
\begin{tabular}{lcccccccc}
\toprule
& \multicolumn{2}{c}{\textbf{IFEval}} & \multicolumn{2}{c}{\textbf{IFBench}} & \multicolumn{1}{c}{\textbf{MulDimIF}} & \multicolumn{2}{c}{\textbf{AdvancedIF}} \\
\cmidrule(lr){2-3}\cmidrule(lr){4-5}\cmidrule(lr){6-6}\cmidrule(lr){7-8}
\textbf{Model / Strategy} & Prompt & Instruct & Prompt & Instruct & Acc. & Overall & Rubric \\
\midrule
Qwen3-4B-Instruct & & & & & & & \\
~~w/ Inter-sample group normalization  & 82.99 & 88.49 & 30.95 & 32.84 & 75.08 & 44.67 & 78.73 \\
\rowcolor{rowblue}
~~\textbf{w/ Intra-sample group normalization}  & \textbf{85.21} & \textbf{88.73} & \textbf{34.69} & \textbf{37.61} & \textbf{76.75} & \textbf{47.48} & \textbf{80.79} \\
\midrule
Qwen2.5-7B-Instruct & & & & & & & \\
~~w/ Inter-sample group normalization  & 82.07 & 86.81 & 28.57 & 29.25 & 69.75 & 31.61 & 72.04 \\
\rowcolor{rowblue}
~~\textbf{w/ Intra-sample group normalization}  & \textbf{82.26} & \textbf{87.29} & \textbf{32.31} & \textbf{33.43} & \textbf{71.83} & \textbf{32.51} & \textbf{72.70} \\
\midrule
Llama3.2-3B-Instruct & & & & & & & \\
~~w/ Inter-sample group normalization  & 77.45 & 84.41 & 24.49 & 25.97 & \textbf{69.42} & 36.05 & 73.49 \\
\rowcolor{rowblue}
~~\textbf{w/ Intra-sample group normalization}  & \textbf{79.48} & \textbf{85.37} & \textbf{26.19} & \textbf{27.46} & 68.58 & \textbf{37.99} & \textbf{74.07} \\
\bottomrule
\end{tabular}%
}
\caption{Ablation on group partitioning strategies across all model configurations. We compare Inter-sample vs.\ Intra-sample Token Group Normalization for advantage estimation. Best results per model are \textbf{bolded}.}
\label{tab:ablation_group}
\end{table}

\subsubsection{A2: Discriminator Weight Strategy}
\label{sec::ablation_discriminator}

To validate the benefit of our proposed Token-level Relevance Discriminator, we compare four token weighting strategies in Table~\ref{tab:ablation_discriminator}: (1) \textit{Random}, where each token is assigned a random weight $w\in \left[0,1\right]$; (2) \textit{Uniform}, where all tokens receive equal weight (equivalent to RL-AON); (3) \textit{Greedy}, where a discriminator trained on data annotated by a naive greedy prompt that directly extracts constraint-related tokens without our structured taxonomy; and (4) \textit{Ours}, the discriminator trained with our full annotation strategy. The results confirm that our annotation strategy provides more reliable token-level supervision than other methods. The prompts of greedy and full annotation strategy are provided in Appendix~\ref{app:annotation_prompts}. We also provide qualitative token-attribution case studies in Section~\ref{sec::analysis_token}.

\begin{table}[h]
\centering
\small
    \begin{tabular}{lccccccc}
    \toprule
    & \multicolumn{2}{c}{\textbf{IFEval}} & \multicolumn{2}{c}{\textbf{IFBench}} & \multicolumn{1}{c}{\textbf{MulDimIF}} & \multicolumn{2}{c}{\textbf{AdvancedIF}} \\
    \cmidrule(lr){2-3}\cmidrule(lr){4-5}\cmidrule(lr){6-6}\cmidrule(lr){7-8}
    \textbf{Strategy} & \multicolumn{1}{c}{Prompt} & \multicolumn{1}{c}{Instruct} & \multicolumn{1}{c}{Prompt} & \multicolumn{1}{c}{Instruct} & \multicolumn{1}{c}{Acc.} & \multicolumn{1}{c}{Overall} & \multicolumn{1}{c}{Rubric} \\
\midrule
\rowcolor{gray!20}\multicolumn{8}{c}{\textit{Base Model: Qwen3-4B-Instruct-2507}} \\
\midrule
Random              & 70.43 & 79.50 & 32.99 & 35.52 & 61.75 & 46.50 & 78.39 \\
Uniform (RL-AON)    & 82.99 & 88.61 & 32.99 & 35.52 & 74.25 & 47.29 & 79.69 \\
Greedy              & 85.03 & 88.49 & 30.95 & 32.84 & 75.08 & 46.20 & 79.82 \\
\rowcolor{rowblue}
\textbf{Ours (RTT-AON)}      & \textbf{85.21} & \textbf{88.73} & \textbf{34.69} & \textbf{37.61} & \textbf{76.75} & \textbf{47.48} & \textbf{80.79} \\
\bottomrule
\end{tabular}
\caption{Ablation on token weighting strategies for RL training.}
\label{tab:ablation_discriminator}
\end{table}

\subsubsection{A3: Sensitivity to the Token-Level Weight ($\beta$)}
\label{sec::ablation_beta}

\begin{wrapfigure}{r}{0.40\textwidth}
\centering
\vspace{-10pt}
\includegraphics[width=0.9\linewidth]{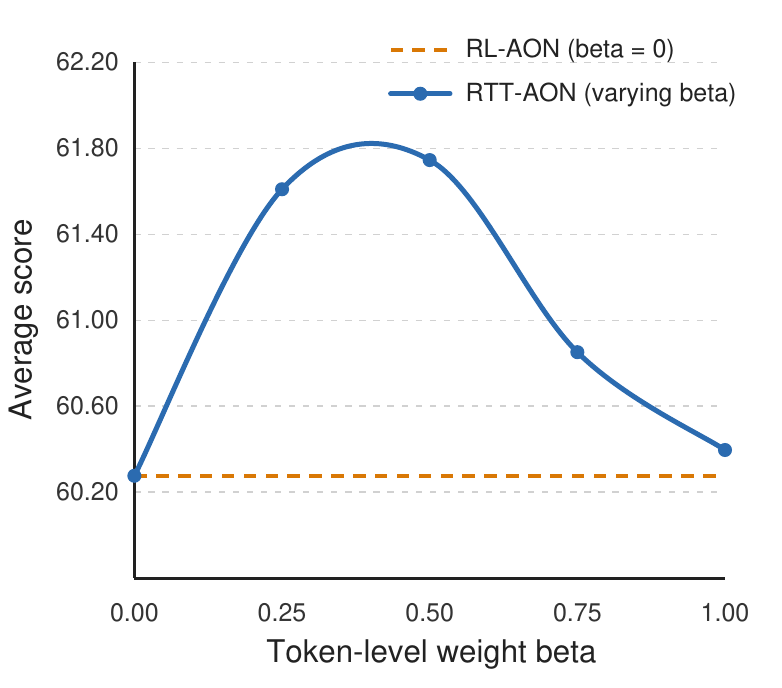}
\caption{Sensitivity of RTT to $\beta$. Baseline $\beta=0$ is shown as a horizontal line.}
\label{fig:beta_sensitivity}
\vspace{-20pt}
\end{wrapfigure}

We examine the sensitivity of RTT to the token-level weight $\beta$ in the joint advantage (Equation~\ref{eq:combined_advantage}). Since the response-level signal is indispensable for distinguishing responses across the rollout group, we fix $\alpha=1$ and vary $\beta$ over $\{0, 0.25, 0.5, 0.75, 1.0\}$ to study the marginal effect of token-level reward. Figure~\ref{fig:beta_sensitivity} reports the average score on IFEval, IFBench, and MulDimIF for RTT-AON and RL-AON trained on Qwen3-4B-Instruct.

According to the results, all nonzero $\beta$ values remain competitive with or above the baseline, confirming that RTT benefits from token-level credit assignment and is robust to the choice of $\beta$. The best performance is achieved at $\beta=0.5$; beyond this point, performance gradually declines. This is because the token-level advantage is normalized \textbf{within} a single response and only captures intra-sample token importance. As $\beta$ grows, this intra-sample signal increasingly dominates the combined advantage, causing the policy to over-correct individual token probabilities and impair overall response quality.

\subsection{Analysis}
\label{sec::analysis}

\subsubsection{Training Dynamics}
\label{sec::analysis_training_dynamics}

\begin{figure}[h]
\centering
\includegraphics[width=\linewidth]{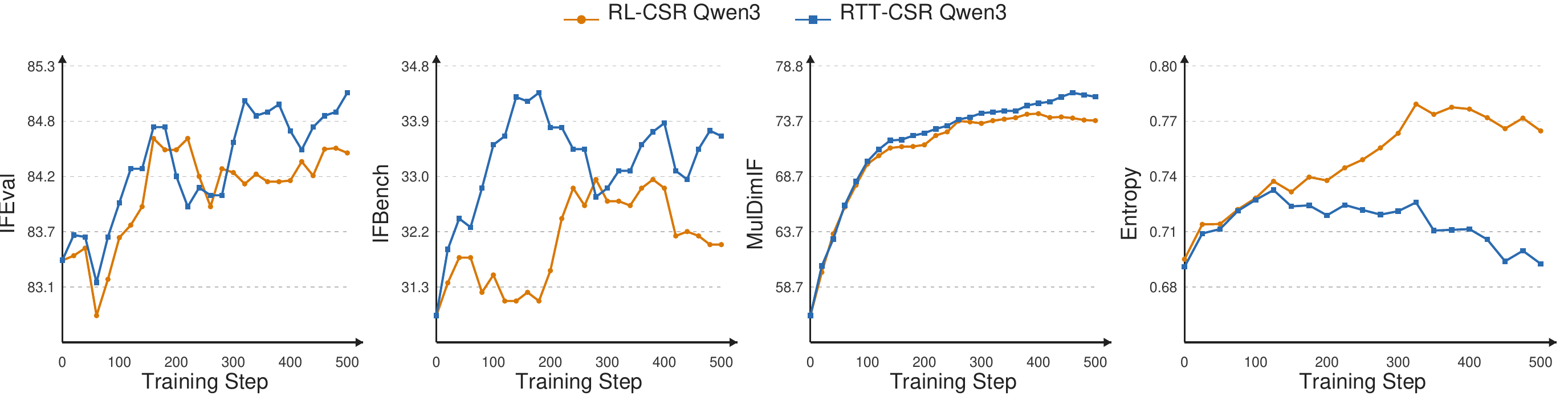}
\caption{Training dynamics of RTT-CSR vs.\ RL-CSR baseline. Left side shows model performance on IFEval, IFBench, and MulDimIF benchmarks, while right side shows response entropy.}
\label{fig:b2_training}
\end{figure}

Figure~\ref{fig:b2_training} compares the training dynamics of RTT-CSR and the RL-CSR. Model performance of RTT-CSR improves steadily across these three benchmarks with higher efficiency than RL-CSR. Meanwhile, RTT-CSR maintains lower response entropy throughout training compared to RL-CSR, with the gap widening over time. This suggests that RTT-CSR token-level credit assignment provides denser and more precise reward signals to guide the policy,  avoiding reward ambiguity problem in RL-CSR training.

\subsubsection{Training Stability on Llama3.2-3B-Instruct}
\label{sec::analysis_llama}

\begin{figure}[h]
\centering
\includegraphics[width=\linewidth]{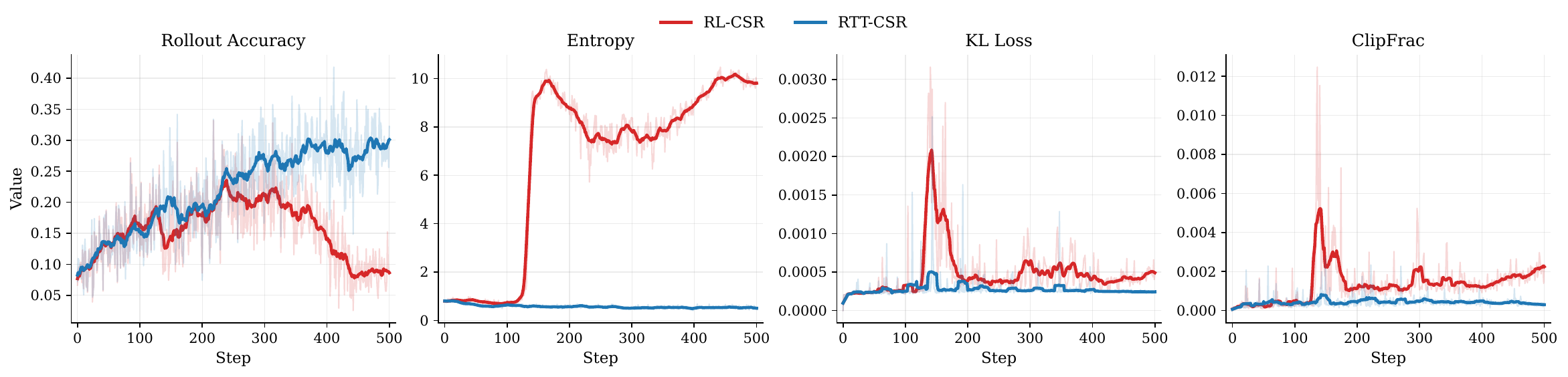}
\caption{Training stability comparison between RTT-CSR and RL-CSR on Llama3.2-3B-Instruct. We report rollout accuracy, entropy, KL loss, and clip fraction over training. RL-CSR exhibits severe instability, while RTT-CSR remains stable.}
\label{fig:llama_training_stability}
\end{figure}

Figure~\ref{fig:llama_training_stability} further analyzes training dynamics on Llama3.2-3B-Instruct through rollout accuracy, entropy, KL loss, and clip fraction. During RL-CSR training, we observe severe instability: rollout accuracy drops sharply, while entropy, KL loss, and clip fraction rise rapidly, leading to poorer performance on IFEval and MulDimIF comparing to RL-AON in Table~\ref{tab:main_results}. By contrast, RTT-CSR does not exhibit these behaviors and remains stable throughout training. This contrast is consistent with the reward ambiguity issue discussed in the Introduction part, suggesting that token-level credit assignment in RTT-CSR can mitigate reward ambiguity and improve training stability.

\subsubsection{Computational Overhead}
\label{sec::analysis_overhead}

\begin{wraptable}{r}{0.5\textwidth}
\centering
\vspace{-12pt}
\small
\begin{tabular}{lrr}
\toprule
\textbf{Training Phase} & \textbf{GPU hours} & \textbf{Fraction} \\
\midrule
Rollout Generation & 1{,}160.7 & 76.8\% \\
Policy Gradient Update & 131.6 & 8.7\% \\
Others (e.g., Communication) & 101.1 & 6.7\% \\
\cmidrule(lr){1-3}
Discriminator Forward & 117.9 & 7.8\% \\
\midrule
\textbf{RTT Total} & \textbf{1{,}511.3} & \textbf{100\%} \\
RL Baseline (w/o Discriminator) & 1{,}393.4 & --- \\
\bottomrule
\end{tabular}
\caption{GPU-hour breakdown for 500-step RL training with Qwen3-4B-Instruct on A100-80GB GPUs. The first three rows constitute the shared RL pipeline; the discriminator forward pass (below the rule) is the only additional cost introduced by RTT.}
\label{tab:overhead}
\vspace{-10pt}
\end{wraptable}

To quantify the additional training cost introduced by the Token-Level Relevance Discriminator, we report the GPU hour breakdown across different phases during the 500-step RTT training of Qwen3-4B-Instruct on NVIDIA A100-80GB GPUs. As shown in Table~\ref{tab:overhead}, rollout generation dominates the computational budget at $76.8\%$, while the forward pass of the discriminator requires only $117.9$ GPU hours ($7.8\%$ of the total), introducing $8.5\%$ additional overhead relative to the RL baseline (i.e., the same pipeline without the discriminator). This low overhead stems from the forward-only nature of the discriminator, which requires neither gradient computation nor optimizer state storage, and consequently incurs far less compute and memory than the policy gradient update phase.

\subsubsection{Token Attribution Case Studies}
\label{sec::analysis_token}

Table~\ref{fig:b1_token_attribution} presents case studies visualizing the token-level relevance scores assigned by Token-level Relevance Discriminator for a given constraint. For a response that satisfies the constraint, the discriminator assigns high relevance scores to the tokens that constitute the compliant content; for a response that violates the constraint, high relevance scores are concentrated around the specific tokens responsible for the violation. This visualization illustrates how RTT localizes the reward signal and resolves the ambiguity inherent in response-level scoring: two responses with the same CSR score can receive different token-level gradients.

\section{Conclusion}
We present Rubrics to Tokens (RTT), a token-level rubric-based reinforcement learning framework that addresses the  limitations of response-level reward in instruction following tasks. By training a Token-Level Relevance Discriminator, RTT projects coarse response-level rubric scores into fine-grained token-level reward signals. We further propose RTT-GRPO, a policy optimization objective that integrates per-constraint token-level advantages with response-level signals within a unified GRPO framework, and the intra-sample token group normalization as a highly effective group partitioning strategy. Extensive experiments demonstrate that RTT consistently outperforms all baselines in both instruction-level and rubric-level accuracy. We hope that RTT could be applied in agentic scenarios, where credit assignment becomes essential for training agents under long-horizon, multi-step constraints.

\begin{table}[!htbp]
\centering
\small
\renewcommand{\arraystretch}{1.2}
\begin{tabular}{p{0.26\linewidth} p{0.73\linewidth}}
\toprule
\textbf{Constraint} & \textbf{Token Attribution} \\
\midrule

\vspace{-5em}\textbf{Make sure your response in all capital letters.}
  & \begin{minipage}[t]{\linewidth}\includegraphics[width=\linewidth]{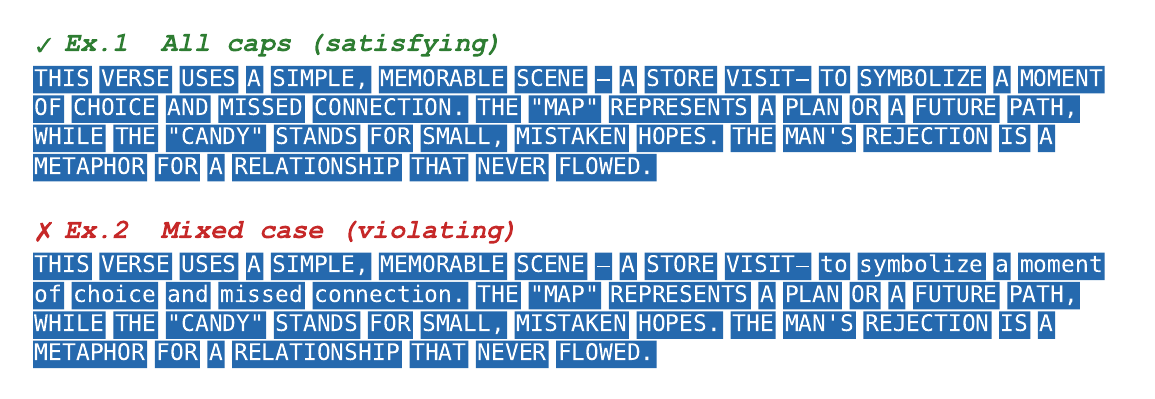}\\[2pt]
    \textit{\textbf{Explanation:} Since the constraint applies globally, the discriminator assigns high relevance to every token in both the satisfying and violating responses. Even when the violating response switches to lowercase mid-sentence, the dense attribution reflects that the entire output is implicated in the constraint.}\end{minipage} \\
\midrule

\vspace{-5em}\textbf{Your response should start with ``My Answer:''}
  & \begin{minipage}[t]{\linewidth}\includegraphics[width=\linewidth]{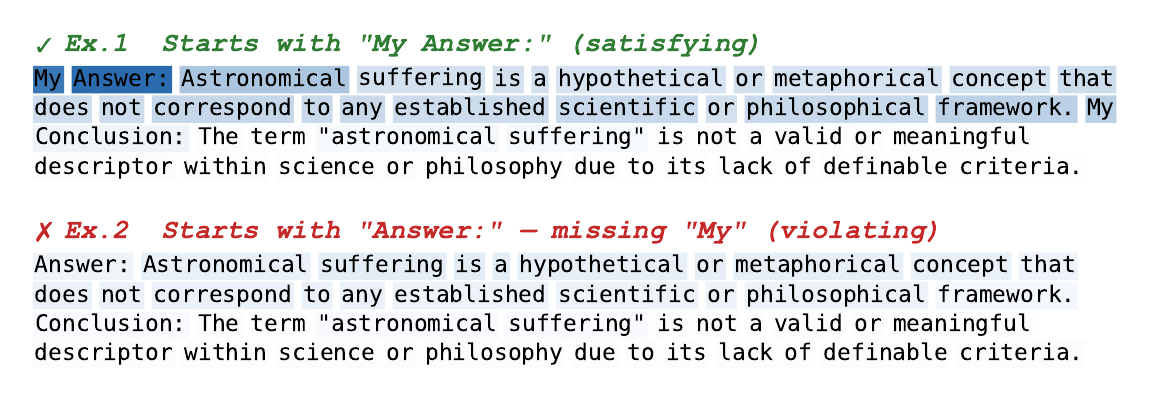}\\[2pt]
    \textit{\textbf{Explanation:} The discriminator concentrates high relevance on the opening tokens ``My'' and ``Answer:'' in the satisfying response. In the violating response, which begins with ``Answer:'' (missing ``My''), relevance scores collapse across the entire output, as the pattern match fails from the first token.}\end{minipage} \\
\midrule

\vspace{-4em}\textbf{Do not include word `ride' in your response.}
  & \begin{minipage}[t]{\linewidth}\includegraphics[width=\linewidth]{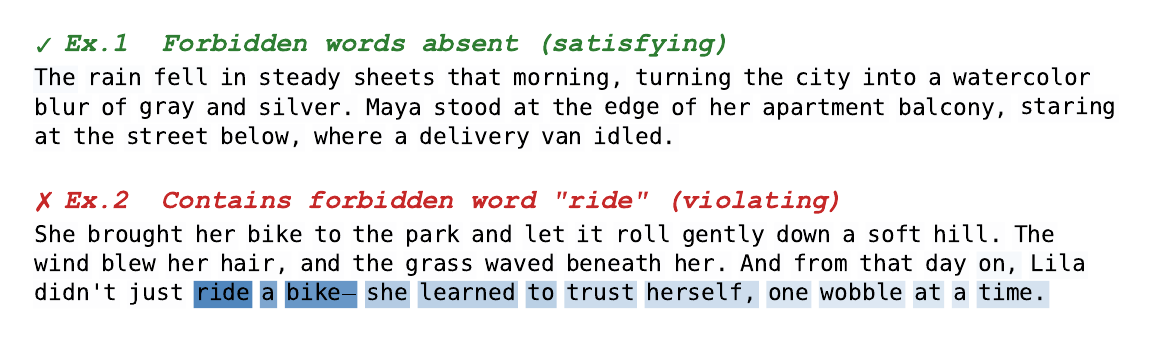}\\[2pt]
    \textit{\textbf{Explanation:} All tokens receive near-zero relevance in the first response, confirming the constraint is met. In the violating response, relevance spikes sharply at the forbidden word ``ride'' and its neighbors, localizing the violation site.}\end{minipage} \\
\midrule

\vspace{-5em}\textbf{State that recent global warming is primarily caused by human greenhouse gas emissions.}
  & \begin{minipage}[t]{\linewidth}\includegraphics[width=\linewidth]{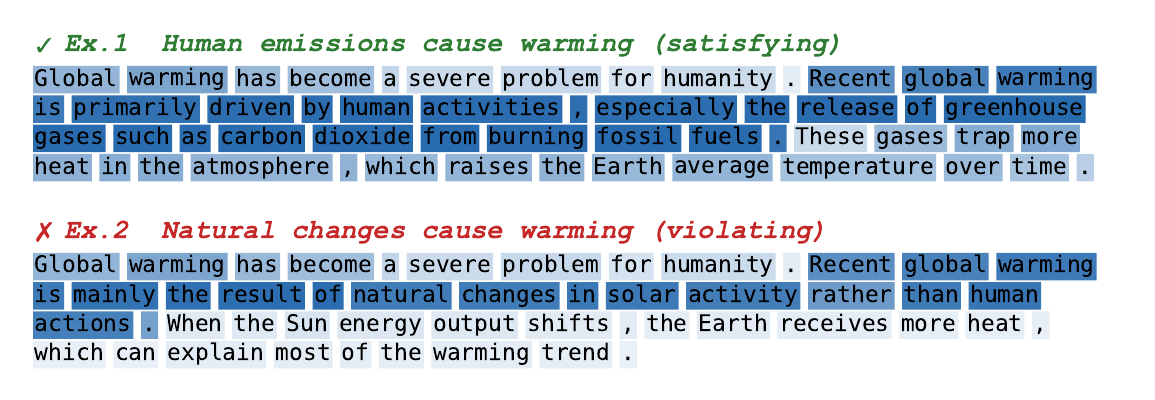}\\[2pt]
    \textit{\textbf{Explanation:} The discriminator also captures semantic content. In the satisfying response, it assigns high relevance to the statement directly expressing the correct claim. In the violating response, relevance concentrates on the misattributing statement, so the corresponding tokens can receive negative reward signals.}\end{minipage} \\

\bottomrule
\end{tabular}
\caption{Token attribution visualization for different constraint types. Darker shading indicates higher discriminator relevance score.}
\label{fig:b1_token_attribution}
\end{table}

\FloatBarrier

\bibliographystyle{related}
\bibliography{related}

@article{srivastava2025technical,
  title={A technical survey of reinforcement learning techniques for large language models},
  author={Srivastava, Saksham Sahai and Aggarwal, Vaneet},
  journal={arXiv preprint arXiv:2507.04136},
  year={2025}
}

@article{zhang2025replay,
  title={Replay failures as successes: Sample-efficient reinforcement learning for instruction following},
  author={Zhang, Kongcheng and Yao, Qi and Liu, Shunyu and Zhang, Wenjian and Cen, Min and Zhou, Yang and Fang, Wenkai and Zhao, Yiru and Lai, Baisheng and Song, Mingli},
  journal={arXiv preprint arXiv:2512.23457},
  year={2025}
}

@inproceedings{peng2025verif,
  title={Verif: Verification engineering for reinforcement learning in instruction following},
  author={Peng, Hao and Qi, Yunjia and Wang, Xiaozhi and Xu, Bin and Hou, Lei and Li, Juanzi},
  booktitle={Proceedings of the 2025 Conference on Empirical Methods in Natural Language Processing},
  pages={30312--30327},
  year={2025}
}

@article{lou2023muffin,
  title={Muffin: Curating multi-faceted instructions for improving instruction-following},
  author={Lou, Renze and Zhang, Kai and Xie, Jian and Sun, Yuxuan and Ahn, Janice and Xu, Hanzi and Su, Yu and Yin, Wenpeng},
  journal={arXiv preprint arXiv:2312.02436},
  year={2023}
}

@article{zhou2023lima,
  title={Lima: Less is more for alignment},
  author={Zhou, Chunting and Liu, Pengfei and Xu, Puxin and Iyer, Srinivasan and Sun, Jiao and Mao, Yuning and Ma, Xuezhe and Efrat, Avia and Yu, Ping and Yu, Lili and others},
  journal={Advances in Neural Information Processing Systems},
  volume={36},
  pages={55006--55021},
  year={2023}
}

@article{xu2023wizardlm,
  title={Wizardlm: Empowering large language models to follow complex instructions},
  author={Xu, Can and Sun, Qingfeng and Zheng, Kai and Geng, Xiubo and Zhao, Pu and Feng, Jiazhan and Tao, Chongyang and Jiang, Daxin},
  journal={arXiv preprint arXiv:2304.12244},
  year={2023}
}

@article{cheng2024spar,
  title={Spar: Self-play with tree-search refinement to improve instruction-following in large language models},
  author={Cheng, Jiale and Liu, Xiao and Wang, Cunxiang and Gu, Xiaotao and Lu, Yida and Zhang, Dan and Dong, Yuxiao and Tang, Jie and Wang, Hongning and Huang, Minlie},
  journal={arXiv preprint arXiv:2412.11605},
  year={2024}
}

@article{liu2025recast,
  title={RECAST: Strengthening LLMs' Complex Instruction Following with Constraint-Verifiable Data},
  author={Liu, Wenhao and Guo, Zhengkang and Xie, Mingchen and Xu, Jingwen and Huang, Zisu and Tian, Muzhao and Xu, Jianhan and Wu, Muling and Wang, Xiaohua and Lv, Changze and others},
  journal={arXiv e-prints},
  pages={arXiv--2505},
  year={2025}
}

@article{dong2024self,
  title={Self-play with execution feedback: Improving instruction-following capabilities of large language models},
  author={Dong, Guanting and Lu, Keming and Li, Chengpeng and Xia, Tingyu and Yu, Bowen and Zhou, Chang and Zhou, Jingren},
  journal={arXiv preprint arXiv:2406.13542},
  year={2024}
}

@article{ouyang2022training,
  title={Training language models to follow instructions with human feedback},
  author={Ouyang, Long and Wu, Jeffrey and Jiang, Xu and Almeida, Diogo and Wainwright, Carroll and Mishkin, Pamela and Zhang, Chong and Agarwal, Sandhini and Slama, Katarina and Ray, Alex and others},
  journal={Advances in neural information processing systems},
  volume={35},
  pages={27730--27744},
  year={2022}
}

@inproceedings{he2024complex,
  title={From complex to simple: Enhancing multi-constraint complex instruction following ability of large language models},
  author={He, Qianyu and Zeng, Jie and He, Qianxi and Liang, Jiaqing and Xiao, Yanghua},
  booktitle={Findings of the Association for Computational Linguistics: EMNLP 2024},
  pages={10864--10882},
  year={2024}
}

@article{cui2025process,
  title={Process reinforcement through implicit rewards},
  author={Cui, Ganqu and Yuan, Lifan and Wang, Zefan and Wang, Hanbin and Zhang, Yuchen and Chen, Jiacheng and Li, Wendi and He, Bingxiang and Fan, Yuchen and Yu, Tianyu and others},
  journal={arXiv preprint arXiv:2502.01456},
  year={2025}
}

@article{guo2023beyond,
  title={Beyond imitation: Leveraging fine-grained quality signals for alignment},
  author={Guo, Geyang and Zhao, Ranchi and Tang, Tianyi and Zhao, Wayne Xin and Wen, Ji-Rong},
  journal={arXiv preprint arXiv:2311.04072},
  year={2023}
}

@article{cao2024beyond,
  title={Beyond sparse rewards: Enhancing reinforcement learning with language model critique in text generation},
  author={Cao, Meng and Shu, Lei and Yu, Lei and Zhu, Yun and Wichers, Nevan and Liu, Yinxiao and Meng, Lei},
  journal={arXiv preprint arXiv:2401.07382},
  year={2024}
}

@inproceedings{yoon2024tlcr,
  title={Tlcr: Token-level continuous reward for fine-grained reinforcement learning from human feedback},
  author={Yoon, Eunseop and Yoon, Hee Suk and Eom, SooHwan and Han, Gunsoo and Nam, Daniel and Jo, Daejin and On, Kyoung-Woon and Hasegawa-Johnson, Mark and Kim, Sungwoong and Yoo, Chang},
  booktitle={Findings of the Association for Computational Linguistics: ACL 2024},
  pages={14969--14981},
  year={2024}
}

@article{tan2025gtpo,
  title={Gtpo and grpo-s: Token and sequence-level reward shaping with policy entropy},
  author={Tan, Hongze and Pan, Jianfei and Lin, Jinghao and Chen, Tao and Zheng, Zhihang and Tang, Zhihao and Yang, Haihua},
  journal={arXiv preprint arXiv:2508.04349},
  year={2025}
}

@article{zhou2025breaking,
  title={Breaking the exploration bottleneck: Rubric-scaffolded reinforcement learning for general llm reasoning},
  author={Zhou, Yang and Li, Sunzhu and Liu, Shunyu and Fang, Wenkai and Zhang, Kongcheng and Zhao, Jiale and Yang, Jingwen and Zhou, Yihe and Lv, Jianwei and Zheng, Tongya and others},
  journal={arXiv preprint arXiv:2508.16949},
  year={2025}
}

@article{gunjal2025rubrics,
  title={Rubrics as rewards: Reinforcement learning beyond verifiable domains},
  author={Gunjal, Anisha and Wang, Anthony and Lau, Elaine and Nath, Vaskar and He, Yunzhong and Liu, Bing and Hendryx, Sean},
  journal={arXiv preprint arXiv:2507.17746},
  year={2025}
}

@article{huang2025reinforcement,
  title={Reinforcement learning with rubric anchors},
  author={Huang, Zenan and Zhuang, Yihong and Lu, Guoshan and Qin, Zeyu and Xu, Haokai and Zhao, Tianyu and Peng, Ru and Hu, Jiaqi and Shen, Zhanming and Hu, Xiaomeng and others},
  journal={arXiv preprint arXiv:2508.12790},
  year={2025}
}

@article{sharma2025researchrubrics,
  title={Researchrubrics: A benchmark of prompts and rubrics for evaluating deep research agents},
  author={Sharma, Manasi and Zhang, Chen Bo Calvin and Bandi, Chaithanya and Wang, Clinton and Aich, Ankit and Nghiem, Huy and Rabbani, Tahseen and Htet, Ye and Jang, Brian and Basu, Sumana and others},
  journal={arXiv preprint arXiv:2511.07685},
  year={2025}
}

@article{goel2025training,
  title={Training AI Co-Scientists Using Rubric Rewards},
  author={Goel, Shashwat and Hazra, Rishi and Jayalath, Dulhan and Willi, Timon and Jain, Parag and Shen, William F and Leontiadis, Ilias and Barbieri, Francesco and Bachrach, Yoram and Geiping, Jonas and others},
  journal={arXiv preprint arXiv:2512.23707},
  year={2025}
}

@article{he2025advancedif,
  title={Advancedif: Rubric-based benchmarking and reinforcement learning for advancing llm instruction following},
  author={He, Yun and Li, Wenzhe and Zhang, Hejia and Li, Songlin and Mandyam, Karishma and Khosla, Sopan and Xiong, Yuanhao and Wang, Nanshu and Peng, Xiaoliang and Li, Beibin and others},
  journal={arXiv preprint arXiv:2511.10507},
  year={2025}
}

@article{wen2024benchmarking,
  title={Benchmarking complex instruction-following with multiple constraints composition},
  author={Wen, Bosi and Ke, Pei and Gu, Xiaotao and Wu, Lindong and Huang, Hao and Zhou, Jinfeng and Li, Wenchuang and Hu, Binxin and Gao, Wendy and Xu, Jiaxing and others},
  journal={Advances in Neural Information Processing Systems},
  volume={37},
  pages={137610--137645},
  year={2024}
}

@article{zhou2023ifeval,
  title={Instruction-following evaluation for large language models},
  author={Zhou, Jeffrey and Lu, Tianjian and Mishra, Swaroop and Brahma, Siddhartha and Basu, Sujoy and Luan, Yi and Zhou, Denny and Hou, Le},
  journal={arXiv preprint arXiv:2311.07911},
  year={2023}
}

@article{pyatkin2025ifbench,
  title={Generalizing verifiable instruction following},
  author={Pyatkin, Valentina and Malik, Saumya and Graf, Victoria and Ivison, Hamish and Huang, Shengyi and Dasigi, Pradeep and Lambert, Nathan and Hajishirzi, Hannaneh},
  journal={arXiv preprint arXiv:2507.02833},
  year={2025}
}

@article{ye2025multi,
  title={A multi-dimensional constraint framework for evaluating and improving instruction following in large language models},
  author={Ye, Junjie and Huang, Caishuang and Chen, Zhuohan and Fu, Wenjie and Yang, Chenyuan and Yang, Leyi and Wu, Yilong and Wang, Peng and Zhou, Meng and Yang, Xiaolong and others},
  journal={arXiv preprint arXiv:2505.07591},
  year={2025}
}

@article{qin2024infobench,
      title={InFoBench: Evaluating Instruction Following Ability in Large Language Models}, 
      author={Yiwei Qin and Kaiqiang Song and Yebowen Hu and Wenlin Yao and Sangwoo Cho and Xiaoyang Wang and Xuansheng Wu and Fei Liu and Pengfei Liu and Dong Yu},
      year={2024},
      eprint={2401.03601},
      archivePrefix={arXiv},
      primaryClass={cs.CL}
}

@article{zheng2023judging,
  title={Judging llm-as-a-judge with mt-bench and chatbot arena},
  author={Zheng, Lianmin and Chiang, Wei-Lin and Sheng, Ying and Zhuang, Siyuan and Wu, Zhanghao and Zhuang, Yonghao and Lin, Zi and Li, Zhuohan and Li, Dacheng and Xing, Eric and others},
  journal={Advances in neural information processing systems},
  volume={36},
  pages={46595--46623},
  year={2023}
}

@article{rafailov2023direct,
  title={Direct preference optimization: Your language model is secretly a reward model},
  author={Rafailov, Rafael and Sharma, Archit and Mitchell, Eric and Manning, Christopher D and Ermon, Stefano and Finn, Chelsea},
  journal={Advances in neural information processing systems},
  volume={36},
  pages={53728--53741},
  year={2023}
}

@inproceedings{zeng2025reinforcing,
  title={Reinforcing multi-turn reasoning in llm agents via turn-level credit assignment},
  author={Zeng, Siliang and Wei, Quan and Brown, William and Frunza, Oana and Nevmyvaka, Yuriy and Zhao, Yang Katie and Hong, Mingyi},
  booktitle={ICML 2025 Workshop on Computer Use Agents},
  year={2025}
}

@article{ping2026longr,
  title={LongR: Unleashing Long-Context Reasoning via Reinforcement Learning with Dense Utility Rewards},
  author={Ping, Bowen and Chen, Zijun and Yu, Yiyao and Hui, Tingfeng and Yan, Junchi and Chang, Baobao},
  journal={arXiv preprint arXiv:2602.05758},
  year={2026}
}

@article{shao2024deepseekmath,
  title={Deepseekmath: Pushing the limits of mathematical reasoning in open language models},
  author={Shao, Zhihong and Wang, Peiyi and Zhu, Qihao and Xu, Runxin and Song, Junxiao and Bi, Xiao and Zhang, Haowei and Zhang, Mingchuan and Li, YK and Wu, Yang and others},
  journal={arXiv preprint arXiv:2402.03300},
  year={2024}
}

@article{lambert2024tulu,
  title={Tulu 3: Pushing frontiers in open language model post-training},
  author={Lambert, Nathan and Morrison, Jacob and Pyatkin, Valentina and Huang, Shengyi and Ivison, Hamish and Brahman, Faeze and Miranda, Lester James V and Liu, Alisa and Dziri, Nouha and Lyu, Shane and others},
  journal={arXiv preprint arXiv:2411.15124},
  year={2024}
}

@inproceedings{lightman2023let,
  title={Let's verify step by step},
  author={Lightman, Hunter and Kosaraju, Vineet and Burda, Yuri and Edwards, Harrison and Baker, Bowen and Lee, Teddy and Leike, Jan and Schulman, John and Sutskever, Ilya and Cobbe, Karl},
  booktitle={The twelfth international conference on learning representations},
  year={2023}
}

@inproceedings{wang2024math,
  title={Math-shepherd: Verify and reinforce llms step-by-step without human annotations},
  author={Wang, Peiyi and Li, Lei and Shao, Zhihong and Xu, Runxin and Dai, Damai and Li, Yifei and Chen, Deli and Wu, Yu and Sui, Zhifang},
  booktitle={Proceedings of the 62nd Annual Meeting of the Association for Computational Linguistics (Volume 1: Long Papers)},
  pages={9426--9439},
  year={2024}
}

@article{zhou2026demystifying,
  title={Demystifying Group Relative Policy Optimization: Its Policy Gradient is a U-Statistic},
  author={Zhou, Hongyi and Ye, Kai and Xu, Erhan and Zhu, Jin and Gong, Shijin and Shi, Chengchun},
  journal={arXiv preprint arXiv:2603.01162},
  year={2026}
}

@article{shrivastava2025sample,
  title={Sample more to think less: Group filtered policy optimization for concise reasoning},
  author={Shrivastava, Vaishnavi and Awadallah, Ahmed and Balachandran, Vidhisha and Garg, Shivam and Behl, Harkirat and Papailiopoulos, Dimitris},
  journal={arXiv preprint arXiv:2508.09726},
  year={2025}
}

@article{gu2025group,
  title={Group causal policy optimization for post-training large language models},
  author={Gu, Ziyin and Wang, Jingyao and Zuo, Ran and Sun, Chuxiong and Song, Zeen and Zheng, Changwen and Qiang, Wenwen},
  journal={arXiv preprint arXiv:2508.05428},
  year={2025}
}

@article{rein2024gpqa,
  title={GPQA: A graduate-level google-proof q\&a benchmark},
  author={Rein, David and Hou, Betty Li and Stickland, Asa Cooper and Petty, Jackson and Pang, Richard Yuanzhe and Dirani, Julien and Michael, Julian and Bowman, Samuel R},
  journal={arXiv preprint arXiv:2311.12022},
  year={2024}
}

@article{wang2024mmluPro,
  title={MMLU-Pro: A more robust and challenging multi-task language understanding benchmark},
  author={Wang, Yubo and Ma, Xueguang and Zhang, Ge and Ni, Yuansheng and Chandra, Abhranil and Guo, Shiguang and Ren, Weiming and Arulraj, Aaran and He, Xuan and Jiang, Ziyan and others},
  journal={arXiv preprint arXiv:2406.01574},
  year={2024}
}

@misc{qwen2025qwen25technicalreport,
      title={Qwen2.5 Technical Report}, 
      author={Qwen and : and An Yang and Baosong Yang and Beichen Zhang and Binyuan Hui and Bo Zheng and Bowen Yu and Chengyuan Li and Dayiheng Liu and Fei Huang and Haoran Wei and others},
      year={2025},
      eprint={2412.15115},
      archivePrefix={arXiv},
      primaryClass={cs.CL},
      url={https://arxiv.org/abs/2412.15115}, 
}

@article{yang2025qwen3,
  title={Qwen3 technical report},
  author={Yang, An and Li, Anfeng and Yang, Baosong and Zhang, Beichen and Hui, Binyuan and Zheng, Bo and Yu, Bowen and Gao, Chang and Huang, Chengen and Lv, Chenxu and others},
  journal={arXiv preprint arXiv:2505.09388},
  year={2025}
}

@article{liu2024deepseek,
  title={Deepseek-v3 technical report},
  author={Liu, Aixin and others},
  journal={arXiv preprint arXiv:2412.19437},
  year={2024}
}

@article{wang2025reinforcement,
  title={Reinforcement Learning Optimization for Large-Scale Learning: An Efficient and User-Friendly Scaling Library},
  author={Wang, Weixun and Xiong, Shaopan and Chen, Gengru and Gao, Wei and Guo, Sheng and He, Yancheng and Huang, Ju and Liu, Jiaheng and Li, Zhendong and Li, Xiaoyang and others},
  journal={arXiv preprint arXiv:2506.06122},
  year={2025}
}

@article{guo2025deepseek,
  title={Deepseek-r1: Incentivizing reasoning capability in llms via reinforcement learning},
  author={Guo, Daya and Yang, Dejian and Zhang, Haowei and Song, Junxiao and Wang, Peiyi and Zhu, Qihao and Xu, Runxin and Zhang, Ruoyu and Ma, Shirong and Bi, Xiao and others},
  journal={arXiv preprint arXiv:2501.12948},
  year={2025}
}

@inproceedings{zheng2024llamafactory,
  title={LlamaFactory: Unified Efficient Fine-Tuning of 100+ Language Models},
  author={Zheng, Yaowei and Zhang, Ruqing and Zhang, Wenhua and Ye, Yong and Luo, Yuan and Yan, Zihan and Zhang, Yujiong and Yang, Yuxuan and Huang, Jing},
  booktitle={Proceedings of the 62nd Annual Meeting of the Association for Computational Linguistics: System Demonstrations},
  year={2024}
}

@article{jiang2025coderl,
  title={CodeRL+: Improving Code Generation via Reinforcement with Execution Semantics Alignment},
  author={Jiang, Xue and Dong, Yihong and Liu, Mengyang and Deng, Hongyi and Wang, Tian and Tao, Yongding and Cao, Rongyu and Li, Binhua and Jin, Zhi and Jiao, Wenpin and others},
  journal={arXiv preprint arXiv:2510.18471},
  year={2025}
}

@article{liu2026gdpo,
  title={GDPO: Group reward-Decoupled Normalization Policy Optimization for Multi-reward RL Optimization},
  author={Liu, Shih-Yang and Dong, Xin and Lu, Ximing and Diao, Shizhe and Belcak, Peter and Liu, Mingjie and Chen, Min-Hung and Yin, Hongxu and Wang, Yu-Chiang Frank and Cheng, Kwang-Ting and Choi, Yejin and Kautz, Jan and Molchanov, Pavlo},
  journal={arXiv preprint arXiv:2601.05242},
  year={2026}
}

@misc{wang2026openclawrltrainagentsimply,
      title={OpenClaw-RL: Train Any Agent Simply by Talking}, 
      author={Yinjie Wang and Xuyang Chen and Xiaolong Jin and Mengdi Wang and Ling Yang},
      year={2026},
      eprint={2603.10165},
      archivePrefix={arXiv},
      primaryClass={cs.CL},
      url={https://arxiv.org/abs/2603.10165}, 
}

@inproceedings{wu2018group,
  title={Group normalization},
  author={Wu, Yuxin and He, Kaiming},
  booktitle={Proceedings of the European conference on computer vision (ECCV)},
  pages={3--19},
  year={2018}
}

@article{hu2024openrlhf,
  title={OpenRLHF: An Easy-to-use, Scalable and High-performance RLHF Framework},
  author={Jian Hu and Xibin Wu and Zilin Zhu and Xianyu and Weixun Wang and Dehao Zhang and Yu Cao},
  journal={arXiv preprint arXiv:2405.11143},
  year={2024}
}

\appendix

\section{Limitations and Future Work}
\label{app:limitations_future}

\subsection{Limitations}

While we identify the Group Partitioning Problem as a fundamental challenge in high-dimensional reward spaces and propose Intra-sample Token Group Normalization as the solution, we only explore two representative strategies (inter-sample vs.\ intra-sample). The three-dimensional reward space admits a richer set of partitioning schemes. For example, \citet{liu2026gdpo} demonstrates that naively normalizing aggregated multi-reward signals can collapse distinct reward combination on multi-reward RL, compressing the training signal. This reward signal collapse resonates with our Group Partitioning Problem: normalization boundaries determine the expressiveness and fidelity of advantage estimates. Exploring other partitioning strategies, such as how the ordering of normalization and aggregation across reward dimensions affects advantage fidelity, remains an open direction.


\subsection{Future Work}

\paragraph{Turn-Level Credit Assignment for Agent Training.}
As LLM-based agents increasingly tackle long-horizon, multi-step tasks through tool use and environment interaction~\citep{wang2026openclawrltrainagentsimply}, credit assignment at the \emph{turn level} becomes a critical bottleneck. In agentic settings, turn-level rewards in a trajectory can be viewed as token-level rewards in a response: the agent must determine which turns (actions, tool calls, or reasoning steps) contributed to eventual task success or failure. RTT's core idea of bridging coarse outcome-level signals with fine-grained rewards naturally extends to this setting, where a Turn-Level Relevance Discriminator identifies which turns satisfy specific sub-goals or constraints. This would enable more precise policy updates in agent RL training, particularly for tasks with sparse, delayed rewards across long interaction trajectories.

\paragraph{Structured Group Partitioning in Multi-Dimensional Reward Spaces.}
The Group Partitioning Problem opens a broader research agenda beyond the two strategies explored here. Future work could investigate other partitioning strategies that partition different dimensions into groups and optimize the policy jointly. As reward spaces grow in dimensionality (e.g., adding sub-task dimensions in compositional reasoning), principled methods for high-dimensional group partitioning will become more important. Connections to group normalization theory~\citep{wu2018group} and multi-objective optimization~\citep{liu2026gdpo} may provide useful theoretical foundations.

\paragraph{Cross-Domain Extension Beyond Instruction Following.}
While RTT is evaluated on instruction-following tasks, its core principle of decomposing holistic evaluation criteria into token-level credit signals is broadly applicable. Promising extensions include long-form generation with multi-faceted quality rubrics (e.g., scientific writing), code generation with multi-dimensional criteria, and creative writing where stylistic and structural constraints serve as independent rubrics. A key open challenge is training a general discriminator that transfers across diverse task types and constraint categories without domain-specific annotation, substantially lowering the barrier to applying RTT in new settings.

\section{Experimental Details}
\label{app:exp_details}

\subsection{Benchmarks and Evaluation Protocols}
\label{app:benchmarks}

\begin{table}[h]
\centering
\small
\begin{tabular}{@{}p{0.22\linewidth}p{0.75\linewidth}@{}}
\toprule
\textbf{Dataset} & \textbf{CoT prompt} \\
\midrule
MATH-500 &
\texttt{Question: \{\}\textbackslash nPlease reason step by step, and put your final answer within \textbackslash boxed\{\}.} \\
\addlinespace[0.4em]
GPQA \& MMLU-Pro &
\texttt{Question: \{\}\textbackslash nAnswer the multiple choice question. The last line of your response should be of the following format: 'Answer: \$LETTER' (without quotes) where LETTER is one of choices. Think step by step before answering.} \\
\bottomrule
\end{tabular}
\caption{CoT evaluation prompts for out-of-domain benchmarks.}
\label{tab:ood_cot_prompts}
\end{table}

\paragraph{Datasets.}
For every benchmark, we evaluate on the full official dataset without subsampling, including in-domain instruction-following benchmarks (IFEval~\citep{zhou2023ifeval}, IFBench~\citep{pyatkin2025ifbench}, MulDimIF~\citep{ye2025multi}, AdvancedIF~\citep{he2025advancedif}) and out-of-domain reasoning benchmarks (MATH-500~\citep{lightman2023let}, GPQA~\citep{rein2024gpqa}, MMLU-Pro~\citep{wang2024mmluPro}).

\paragraph{Metrics and table columns.}
Table columns follow the two granularities defined in the main text (instruction-level vs.\ rubric-level), but column names differ by benchmark:
\begin{itemize}
    \item \textbf{IFEval and IFBench.} The \textit{Prompt} column reports instruction-level accuracy (AON pass rate over prompts). The \textit{Instruct} column reports rubric-level accuracy (average constraint satisfaction rate, CSR). Both benchmarks are reported in the strict mode.
    \item \textbf{AdvancedIF.} The \textit{Overall} column is instruction-level accuracy; while the \textit{Rubric} column is rubric-level accuracy.  Following the official evaluation script, we choose \textbf{o3-mini} as the judge model.
    \item \textbf{MulDimIF.} We report instruction-level accuracy only (single \textit{Acc.} column), consistent with the standard MulDimIF reporting setup.
\end{itemize}

\paragraph{Decoding.} We follow the evaluation recipe in the \textbf{Qwen3 Tech Report}~\citep{yang2025qwen3} for generation: sampling with temperature $0.7$, top-$p$ $0.8$, and top-$k$ $20$. We generate five independent completions per instance and report the averaged score across all benchmarks.

\paragraph{Prompt templates.}
For instruction-following tasks, we use each model's default prompt template (chat template / tokenizer formatting) at evaluation time. For OOD tasks (i.e., MATH-500, GPQA, and MMLU-Pro), we add the following prompts in Table~\ref{tab:ood_cot_prompts}.

\subsection{Training and Evaluation Hyperparameters}
\label{app:hyperparams}

We use LLaMA-Factory~\citep{zheng2024llamafactory} for SFT and DPO training, ROLL~\citep{wang2025reinforcement} for RL training, and OpenRLHF~\citep{hu2024openrlhf} for discriminator training. The detailed training configurations of SFT and DPO are provided in Table~\ref{tab:hyperparams_training}\,(a), the training configurations of RL are provided in Table~\ref{tab:hyperparams_training}\,(b), and training configurations of the discriminator are provided in Table~\ref{tab:hyperparams_training}\,(c). Experiments are conducted on NVIDIA A100-80GB GPUs.

\begin{table}[t]
\centering
\small
\setlength{\tabcolsep}{8pt}
\begin{tabular}{@{}>{\raggedright\arraybackslash}p{0.22\linewidth} @{\hspace{6pt}} L@{}}
\multicolumn{2}{@{}l@{}}{\textit{(a) Training configuration of SFT and DPO}} \\
\toprule
\textbf{Method} & SFT \& DPO \\
\midrule
\textbf{Training} & per\_device\_train\_batch\_size = 4 \& 2, \newline
gradient\_accumulation\_steps = 64 \& 128, \newline
learning\_rate = $10^{-6}$, lr\_scheduler\_type = constant,\newline
cutoff\_len = 2048, warmup\_steps = 10, epochs = 4 \\
\midrule
\textbf{Optimizations} & deepspeed: z2 \& z3, bf16 \\
\bottomrule
\end{tabular}

\vspace{1.2em}

\begin{tabular}{@{}>{\raggedright\arraybackslash}p{0.22\linewidth} @{\hspace{6pt}} L@{}}
\multicolumn{2}{@{}l@{}}{\textit{(b) Training configuration of RL}} \\
\toprule
\textbf{Method} & RL-AON, RL-CSR, RTT \\
\midrule
\textbf{Inference} & top\_k = 100, top\_p = 0.99, temperature = 0.99, num\_return\_sequences\_in\_group = 8,\newline
prompt\_length = 2{,}048, response\_length = 4{,}096 \\
\midrule
\textbf{Training} & per\_device\_train\_batch\_size = 4, gradient\_accumulation\_steps = 2,\newline
rollout\_batch\_size = 64,
learning\_rate = $10^{-6}$, max\_steps = 500. \newline
For Qwen, use\_pg\_clip\_range=true, pg\_clip\_low=0.2, pg\_clip\_high=0.27; \newline
for Llama, use\_pg\_clip\_range=false, pg\_clip=0.2 \\
\midrule
\textbf{Discriminator} &  flash\_attn\_2, bf16 \\
\bottomrule
\end{tabular}

\vspace{1.2em}

\begin{tabular}{@{}>{\raggedright\arraybackslash}p{0.22\linewidth} @{\hspace{6pt}} L@{}}
\multicolumn{2}{@{}l@{}}{\textit{(c) Training configuration of Discriminator}} \\
\toprule
\textbf{Backbone} & Qwen3-4B-Base, Llama3.2-3B-Base \\
\midrule
\textbf{Training} & micro\_train\_batch\_size = 4, train\_batch\_size = 256,\newline
learning\_rate = $9 \times 10^{-6}$, max\_epochs = 4, max\_len = 4{,}096 \\
\midrule
\textbf{Optimizations} & zero\_stage = 2, bf16 \\
\bottomrule
\end{tabular}
\caption{Training configurations across methods and model backbones.}
\label{tab:hyperparams_training}
\end{table}

\subsection{LLM-as-a-Judge for Soft Constraints}
\label{app:judge_prompt}

We evaluate $\mathcal{C}_{\text{soft}}$ with DeepSeek-V3~\citep{liu2024deepseek} using the prompt template below.

\begin{tcolorbox}[
    colback=green!5,
    colframe=green!40!black,
    title={Judge prompt for soft constraints},
    fonttitle=\bfseries\color{white},
    colbacktitle=green!40!black,
    coltitle=white,
    rounded corners,
    breakable,
    enhanced
]
\small
Based on the provided Input (if any) and Generated Text, judge whether the generated text fulfills the Criteria Item with either a YES or NO choice. Your selection should be based on your judgment as well as the following rules:
\newline

- \textbf{YES}: Select ``YES'' if the generated text entirely fulfills the condition specified in the Criteria Item. However, note that even minor inaccuracies exclude the text from receiving a `YES' rating. As an illustration, consider a Criteria Item ``Each sentence in the generated text uses a second person''. If even one sentence does not use the second person, the answer should NOT be `YES'. To qualify for a `YES' rating, the generated text must be entirely accurate and satisfy the criteria.
\newline

-\textbf{NO}: Opt for ``NO'' if the generated text fails to meet the criteria or provides no information that could be utilized to judge. For instance, the Criteria Item asks ``Is the second sentence in the generated text a compound sentence?'' and the generated text only has one sentence. It offers no relevant information to judge whether this criteria is met. Consequently, the answer should be ``NO''.
\newline

\textbf{Input:} \{prompt\}

\textbf{Generated Text:} \{response\}

\textbf{Criteria Item:} \{rubric\}
\newline

You only need to judge whether the generated text satisfies the given Criteria Item and do NOT affect by other requirements in Input (if any). Return either a ``YES'' or ``NO'' choice without any additional text in your response.
\end{tcolorbox}

\section{Discriminator Data and Annotation}
\label{app:discriminator_annotation}

\subsection{Discriminator Prompt}
\label{app:infer_prompt}

We use the following prompt as the discriminator input during training and inference:

\begin{tcolorbox}[
    rttpromptbox,
    colback=gray!5,
    colframe=gray!55!black,
    colbacktitle=gray!70!black,
    title={Prompt template for token-level relevance discriminator},
]
\small\ttfamily
Identify which tokens in the response are relevant to the criteria.\par
\medskip
Criteria: \{criteria\}\par
\medskip
Response: \{response\}\par
\end{tcolorbox}

\subsection{Data Sources and Construction}
\label{app:data_sources}

We build the discriminator corpus in two stages, first constructing an initial instruction-response dataset, then augmenting it with negative samples. To maximize stylistic and structural diversity, we employ multiple frontier LLMs (e.g. Gemini-3-Pro, GPT-5.2, and DeepSeek-V3) to generate diverse responses. 

Specifically, we first draw 22{,}656 examples from the Tulu-3 SFT corpus~\citep{lambert2024tulu}, and all 16{,}969 samples from HiR-16K~\citep{zhang2025replay}. instructions in HiR-16K are further paired with responses generated by Gemini-3-Pro and DeepSeek-V3. 

We then draw instructions from this initial dataset randomly, and synthesize negative responses with two strategies: \emph{minimal modification} and \emph{constraint omission}. In our pipeline, minimal modification is executed with GPT-5.2, while constraint omission uses Qwen3-4B-Instruct. After merging positives and these negatives, the full training set comprises 62{,}714 samples used for token-level annotation and training.

\subsection{Annotation Taxonomy}
\label{app:annotation_taxonomy}

For each constraint-response pair, we categorize constraints into two primary scopes:

\paragraph{Global Scope} The constraint applies holistically to the entire response (e.g., overall tone, count or frequency limits, or formatting structure). In this case, every token in the response is deemed relevant (marked as \texttt{all\_relevant}).

\paragraph{Local Scope}  The constraint strictly concerns specific elements or spans within the response (e.g., the inclusion of a specific keyword, or the absence of a certain phrase). Local constraints are further subdivided based on frequency and polarity:
\begin{itemize}
    \item \textbf{Positive Polarity:} The constraint requires a specific element to be present. If the requirement is satisfied, the LLM annotator extracts the exact verbatim text segments fulfilling the constraint (marked as \texttt{partial\_relevant}). If it is not satisfied, the response is marked as \texttt{all\_irrelevant}.
    \item \textbf{Negative Polarity:} The constraint requires an element to be absent. If the response violates this constraint, the specific violating segments are extracted (marked as \texttt{partial\_relevant}). If no violation occurs (the constraint is satisfied), the response is marked as \texttt{all\_irrelevant}.
\end{itemize}



Once the relevant verbatim text segments are extracted by the LLM annotator, we use the model's tokenizer to map these substrings back to the exact token indices in the response $y$. Tokens falling within these relevant segments are assigned a binary label $l_t = 1$, while other tokens receive label $l_t = 0$. For the \texttt{all\_relevant} cases, all tokens are labeled $l_t = 1$. For the \texttt{all\_irrelevant} cases, we set $l_t = 0$ for all tokens.

\subsection{Annotation Prompts}
\label{app:annotation_prompts}

\paragraph{Response generation.} For the normal completion and constraint omission strategy, we use the following prompt template to generate corresponding responses (note that for constraint omission, we drop one constraint from rubric):

\begin{tcolorbox}[
    rttpromptbox,
    colback=cyan!5,
    colframe=cyan!55!black,
    colbacktitle=cyan!70!black,
    title={Prompt for completion and constraint omission},
]
\small

\textbf{Context \& Task:} Please answer the following Question with a detailed explanation. Your response should be a single, cohesive piece of writing that logically leads the reader from the problem to the solution.
\newline

\textbf{Inputs:}

- \textbf{Question:} \{question\}

- \textbf{Rubric:} \{Rubric\}
\newline

\textbf{Requirements:}
\begin{enumerate}
    \item \textbf{Integration:} Do not separate the reasoning from the answer. Instead, provide a thorough narrative that explains the "how" and "why" alongside the final result.
    \item \textbf{Rubric Fulfillment:} Every constraint listed in the Rubrics must be strictly followed within the flow of your detailed response.
    \item \textbf{Depth:} Avoid brief or "final-answer-only" responses. Elaborate on the logic and steps required to reach the conclusion.
    \newline
\end{enumerate}

\textbf{Output:} \{OUTPUT\}

\end{tcolorbox}

For the minimal modification strategy, we use the following prompt to generated negative response that violates the specific constraint.

\begin{tcolorbox}[
    rttpromptbox,
    colback=teal!6,
    colframe=teal!55!black,
    colbacktitle=teal!65!black,
    title={Prompt for Minimal modification},
]
\small

You are given a question, an original response that follows a specific criterion.
\newline

\textbf{Question:} \{question\}

\textbf{Criterion:} \{constraint\}

\textbf{Original response:} \{original\_response\}
\newline

\textbf{Task:} Rewrite the response with MINIMAL edits so that it VIOLATES this criterion. Keep the rest of the content, style, and length as unchanged as possible. Only make the smallest change necessary to break this single rule. Output ONLY the rewritten response text, no explanation.

\end{tcolorbox}

\paragraph{Token-level annotation.}
The LLM annotator uses a structured prompt covering GLOBAL vs.\ LOCAL branches and positive vs.\ negative polarity. The \textit{Greedy} baseline in Table~\ref{tab:ablation_discriminator} instead uses a single span-extraction template without the taxonomy.

\begin{tcolorbox}[
    rttpromptbox,
    colback=violet!6,
    colframe=violet!45!black,
    colbacktitle=violet!55!black,
    title={Full annotation prompt},
]
\small

You are a training data annotator for a token-level discriminator model. Your task is to classify the Criteria and label which tokens in the Response are relevant to it.
\newline

\textbf{Criteria:}
\{criteria\}

\textbf{Response:}
\{response\}
\newline

Please follow these steps:

\textbf{Step 1: Scope Classification}
\newline

Determine whether this criteria is GLOBAL or LOCAL:

- \textbf{GLOBAL:} The criteria applies to ALL tokens of the response as a whole (e.g., language/writing style requirement, tone requirement, word count limit, paragraph count limit, overall format/structure requirement, topic requirement, count, frequency, or numeric threshold)

- \textbf{LOCAL:} The criteria only concerns SPECIFIC parts of the response (e.g., a specific phrase must appear, a specific section must exist, certain keywords must/must not be used, a particular element must be included/excluded)
\newline

\textbf{Step 2: Polarity Classification}
\newline

If criteria is LOCAL, then determine whether this criteria is POSITIVE or NEGATIVE:

- \textbf{POSITIVE:} Requires something to be present or done (e.g., "must include X", "should start with Y", "the first paragraph should be X")

- \textbf{NEGATIVE:} Requires something to be absent or avoided (e.g., "do not use X", "must not include Y", "exclude the word Z", "avoid mentioning W")
\newline

\textbf{Step 3: Output}
\newline

- \textbf{If GLOBAL}, then output: \{``type": ``all\_relevant''\}

- \textbf{If LOCAL + POSITIVE:}
\begin{itemize}
    \item If relevant text segments EXIST in the response (i.e., the criteria IS satisfied by some portion of the text), then output: \{``type": "partial\_relevant", ``relevant\_texts": [``exact segment 1", ``exact segment 2", ...]\}
    \item If NO relevant text segments exist (i.e., the criteria is NOT satisfied at all), then output: \{"type": "all\_irrelevant"\}
\end{itemize}

- \textbf{If LOCAL + NEGATIVE:}
\begin{itemize}
    \item If text segments that VIOLATE the criteria EXIST in the response, then output: \{``type": ``partial\_relevant", ``relevant\_texts": [``exact violating segment 1", ...]\}
    \item If NO violating text exists (i.e., the criteria IS satisfied / not violated), then output: \{"type": "all\_irrelevant"\}
    \newline
\end{itemize}

\textbf{Important Notes:}
\begin{enumerate}
    \item The text segments in relevant\_texts must be exact original text from the Response (exact match, including punctuation and spaces).
    \item Each text segment should be a continuous, complete sentence or phrase.
    \item Output only JSON format, without any markdown code block markers or other text.
    \item Ensure the JSON format is correct and can be directly parsed.
\end{enumerate}

\end{tcolorbox}

\begin{tcolorbox}[
    rttpromptbox,
    colback=orange!5,
    colframe=orange!65!black,
    colbacktitle=orange!75!black,
    title={Greedy baseline prompt},
]
\small

You are a professional text analysis assistant. Your task is to extract text segments from the Response that are relevant to the given Criteria.
\newline

\textbf{Criteria:} 
\{criteria\}

\textbf{Response:} 
\{response\}
\newline

Please carefully analyze the Response and extract the text segments that address or relate to the Criteria. Even if the entire response seems broadly related, you should identify and extract the MOST RELEVANT and SPECIFIC segments that directly match the Criteria.
\newline

\textbf{Key Principles:}
\begin{itemize}
    \item Just extract the most relevant segments instead of the entire response.
    \item Focus on segments that specifically address the Criteria, not general context or background information.
    \item Be precise: exclude indirect references, examples that don't directly relate, or supporting information that doesn't directly address the Criteria.
    \item If no segments are directly relevant, indicate that the response is ``all\_irrelevant''
    \item If you are sure that all contents are relevant, indicate that the response is ``all\_relevant''
    \newline
\end{itemize}

Please strictly follow the JSON format below and do not add any other text:

1. If the entire response is relevant to the Criteria, output: \{"type": "all\_relevant"\}

2. If the entire response is irrelevant to the Criteria, output: \{"type": "all\_irrelevant"\}

3. If partially relevant, output: \{"type": "partial\_relevant", "relevant\_texts": ["relevant text segment 1", "relevant text segment 2", ...]\}
\newline

\textbf{Important Notes:}
\begin{enumerate}
    \item The text segments in relevant\_texts must be exact original text from the Response (exact match, including punctuation and spaces).
    \item Each text segment should be a continuous, complete sentence or phrase.
    \item Output only JSON format, without any markdown code block markers or other text.
    \item Ensure the JSON format is correct and can be directly parsed.
\end{enumerate}

\end{tcolorbox}

\section{Inter-Sample Token Group Normalization}
\label{app:inter_sample_norm}

We first provide a formal definition of \textbf{Inter-Sample Token Group Normalization}, the baseline strategy discussed in Section~\ref{sec::grpo_group}. Then we give an elaborate analysis of length-bias problem to formally demonstrate why this group normalization method systematically suppresses the advantage signals of shorter responses.

\paragraph{Formal Definition.} Given a constraint $c_k \in \mathcal{C}$ and a group of $G$ sampled responses $\{o_1, \dots, o_G\}$ for instruction $q$, where response $o_i$ has length $T_i$, the token-level discriminator produces relevance probabilities for each response, denoted as $\smash{P^{(i)} = \bigl\{p^{(i,t)}_{c_k} \;\big|\; t = 1, \dots, T_i\bigr\}}$. The token-level reward for $o_i$ under constraint $c_k$ is defined as $\smash{r^{(i,t)}_{c_k} = \mathrm{Score}(o_i, c_k) \cdot p^{(i,t)}_{c_k}}$.

In Inter-Sample Token Group Normalization, \textbf{the normalization group is defined as the all tokens across \emph{all} $G$ responses, for each constraint $c_k$ separately.} Let $\smash{N = \sum_{i=1}^{G} T_i}$ denote the total number of tokens in the group. The group mean and standard deviation are:
\begin{equation}
    \mu_{c_k} = \frac{1}{N} \sum_{i=1}^{G} \sum_{t=1}^{T_i} r^{(i,t)}_{c_k}, \qquad
    \sigma_{c_k} = \sqrt{\frac{1}{N} \sum_{i=1}^{G} \sum_{t=1}^{T_i} \bigl(r^{(i,t)}_{c_k} - \mu_{c_k}\bigr)^2}.
\end{equation}
The per-constraint token-level advantage is then computed as ${A_{c_k}^{(i,t)} = {(r^{(i,t)}_{c_k} - \mu_{c_k})}/{\sigma_{c_k}}}$. And the final token-level advantage is averaged across all constraints: $\hat{A}_{\mathrm{tok}}^{(i,t)} = \frac{1}{|\mathcal{C}|} \sum_{c_k \in \mathcal{C}} A_{c_k}^{(i,t)}$.

\paragraph{Length-bias analysis.} First we define the per-response mean and variance for response $o_i$ under constraint $c_k$ as:
\begin{equation}
    \bar{r}_i = \frac{1}{T_i}\sum_{t=1}^{T_i} r^{(i,t)}_{c_k}, \qquad
    s_i^2 = \frac{1}{T_i}\sum_{t=1}^{T_i}\bigl(r^{(i,t)}_{c_k} - \bar{r}_i\bigr)^2.
\end{equation}
Since $N = \sum_{i=1}^{G} T_i$, the inter-sample mean decomposes exactly as:
\begin{equation}
    \mu_{c_k}
    = \frac{1}{N}\sum_{i=1}^{G}\sum_{t=1}^{T_i} r^{(i,t)}_{c_k}
    = \sum_{i=1}^{G} \frac{T_i}{N}\,\bar{r}_i
    = \sum_{i=1}^{G} w_i\,\bar{r}_i,
    \quad \text{where} ~w_i = \frac{T_i}{N},~ \sum_{i=1}^G w_i = 1.
\end{equation}
Thus $\mu_{c_k}$ is a \emph{length-weighted} average of the per-response means, with weight $w_i \propto T_i$ assigned to response $o_i$. Consequently, longer responses contribute more to the global centering statistic.

Fix the shortest response $o_j$, and let $w_j = T_j / N$ with $T_j \ll T_i$ for all $i \neq j$. Define the leave-one-response-out mean by
\begin{equation}
    \mu_{c_k}^{(-j)}
    = \sum_{i \neq j} \frac{T_i}{N-T_j}\,\bar{r}_i.
\end{equation}
The global mean then satisfies the exact decomposition
\begin{equation}
    \mu_{c_k}
    = w_j \bar{r}_j + (1-w_j)\mu_{c_k}^{(-j)}
    = \mu_{c_k}^{(-j)} + w_j\bigl(\bar{r}_j - \mu_{c_k}^{(-j)}\bigr).
\end{equation}
It follows that the contribution of $o_j$ to the centering term is linear in $w_j$. In the limit $w_j \to 0$, the mean $\mu_{c_k}$ converges to the leave-one-out mean $\smash{\mu_{c_k}^{(-j)}}$ induced by the remaining responses.

By the law of total variance, the inter-sample variance admits the decomposition
\begin{equation}
    \sigma_{c_k}^2
    = \underbrace{w_j s_j^2 + \sum_{i\ne j} w_i s_i^2}_{\text{within-response variance}}
    +\underbrace{w_j\bigl(\bar{r}_j - \mu_{c_k}\bigr)^2 
    + \sum_{i\ne j}w_i\bigl(\bar{r}_i - \mu_{c_k}\bigr)^2}_{\text{between-response variance}}.
\end{equation}

Let $\smash{\bigl(\sigma_{c_k}^{(-j)}\bigr)^2}$ denote the variance computed from the remaining $G$-1 responses:
\begin{equation}
    \bigl(\sigma_{c_k}^{(-j)}\bigr)^2
    = \sum_{i \neq j} \frac{T_i}{N-T_j}\,s_i^2
    + \sum_{i \neq j} \frac{T_i}{N-T_j}\,\bigl(\bar{r}_i - \mu_{c_k}^{(-j)}\bigr)^2.
\end{equation}
Applying the variance formula for a two-component mixture, we can derive:
\begin{equation}
    \begin{aligned}
    \sigma_{c_k}^2
    &= w_j s_j^2 + \sum_{i\ne j} w_i s_i^2 + w_j\bigl(\bar{r}_j - \mu_{c_k}\bigr)^2 
    + \sum_{i\ne j}w_i\bigl(\bar{r}_i - \mu_{c_k}\bigr)^2\\
     &= w_j s_j^2 + (1-w_j)(\sigma_{c_k}^{(-j)})^2 - \sum_{i\ne j}w_i\bigl(\bar{r}_i - \mu_{c_k}^{(-j)}\bigr)^2+ w_j\bigl(\bar{r}_j - \mu_{c_k}\bigr)^2 + \sum_{i\ne j}w_i\bigl(\bar{r}_i - \mu_{c_k}\bigr)^2\\
    \end{aligned}
    \label{eq:sig_c_k}
\end{equation}

For item $\smash{\sum_{i\ne j}w_i\bigl(\bar{r}_i - \mu_{c_k}\bigr)^2}$, we can derive:
\begin{equation}
        \small
    \begin{aligned}
    \sum_{i\ne j}w_i\bigl(\bar{r}_i - \mu_{c_k}\bigr)^2 
    &= \sum_{i\ne j}w_i\bigl(\bar{r}_i - \mu_{c_k}^{(-j)}+\mu_{c_k}^{(-j)}-\mu_{c_k}\bigr)^2\\
    &= \sum_{i\ne j}w_i\bigl(\bar{r}_i - \mu_{c_k}^{(-j)}\bigr)^2 + 2 \bigl(\mu_{c_k}^{(-j)}-\mu_{c_k}\bigr)\sum_{i\ne j}w_i\bigl(\bar{r}_i - \mu_{c_k}^{(-j)}\bigr) + \sum_{i\ne j} w_i\bigl(\mu_{c_k}^{(-j)}-\mu_{c_k}\bigr)^2\\
    &= \sum_{i\ne j}w_i\bigl(\bar{r}_i - \mu_{c_k}^{(-j)}\bigr)^2 + 2 \bigl(\mu_{c_k}^{(-j)}-\mu_{c_k}\bigr)\bigl(\sum_{i\ne j}w_i\bar{r}_i - \sum_{i\ne j}w_i\mu_{c_k}^{(-j)}\bigr) + \sum_{i\ne j} w_i\bigl(\mu_{c_k}^{(-j)}-\mu_{c_k}\bigr)^2\\
    \end{aligned}
\end{equation}

As $\sum_{i\ne j}w_i\bar{r}_i = (1-w_j)\mu_{c_k}^{(-j)}=\mu_{c_k}^{(-j)}\sum_{i\ne j}w_i$, 

we have: $\sum_{i\ne j}w_i\bigl(\bar{r}_i - \mu_{c_k}\bigr)^2 = \sum_{i\ne j}w_i\bigl(\bar{r}_i - \mu_{c_k}^{(-j)}\bigr)^2 + (1-w_j)\bigl(\mu_{c_k}^{(-j)}-\mu_{c_k}\bigr)^2$. 

Substituting back into equation~(\ref{eq:sig_c_k}), we obtain:
\begin{equation}
    \small
    \begin{aligned}
    \sigma_{c_k}^2
     &= w_j s_j^2 + (1-w_j)(\sigma_{c_k}^{(-j)})^2 - \smash\sum_{i\ne j}w_i\bigl(\bar{r}_i - \mu_{c_k}^{(-j)}\bigr)^2+ w_j\bigl(\bar{r}_j - \mu_{c_k}\bigr)^2 + \smash\sum_{i\ne j}w_i\bigl(\bar{r}_i - \mu_{c_k}\bigr)^2\\
     &= w_j s_j^2 + (1-w_j)(\sigma_{c_k}^{(-j)})^2+ w_j\bigl(\bar{r}_j - \mu_{c_k}\bigr)^2 + (1-w_j)\bigl(\mu_{c_k}^{(-j)}-\mu_{c_k}\bigr)^2\\
     &=  w_j s_j^2 + (1-w_j)\bigl(\sigma_{c_k}^{(-j)}\bigr)^2 + w_j(1-w_j)\bigl(\bar{r}_j - \mu_{c_k}^{(-j)}\bigr)^2. 
    \end{aligned}
\end{equation}

Hence the contribution of $o_j$ to the scaling term is likewise controlled by its length weight $w_j$. As $w_j \to 0$, one obtains $\smash{\sigma_{c_k}^2 \to \bigl(\sigma_{c_k}^{(-j)}\bigr)^2}$, so both normalization statistics are asymptotically determined by the other responses.

Since $\smash{\mu_{c_k} \to \mu_{c_k}^{(-j)}}$ and $\smash{\sigma_{c_k} \to \sigma_{c_k}^{(-j)}}$ as $\smash{w_j \to 0}$ (assuming $\smash{\sigma_{c_k}^{(-j)} > 0}$), the normalized token-level advantage for response $o_j$ is
\begin{equation}
    \lim_{w_j \to 0} A_{c_k}^{(j,t)}
    = \lim_{w_j \to 0} \frac{r^{(j,t)}_{c_k} - \mu_{c_k}}{\sigma_{c_k}}
    = \frac{r^{(j,t)}_{c_k} - \mu_{c_k}^{(-j)}}{\sigma_{c_k}^{(-j)}}.
\end{equation}

Hence, the token-level advantages of the shorter response are normalized by statistics induced by the remaining responses. The centering and scaling of $o_j$ are asymptotically governed primarily by external group statistics rather than by statistics specific to $o_j$ itself. This is the source of length bias in inter-sample token group normalization.

By contrast, \textbf{Intra-Sample Token Group Normalization} (Section~\ref{sec::token_advantage}) computes $\mu_R$ and $\sigma_R$ within each response independently. The associated normalization statistics therefore depend only on its own token-level rewards, eliminating the length-bias effect inherent in inter-sample group normalization.

\end{document}